\documentclass[10pt,twocolumn,letterpaper]{article}

\usepackage{iccv}
\usepackage{times}
\usepackage{epsfig}
\usepackage{graphicx}
\usepackage{grffile}
\usepackage{amsmath}
\usepackage{amssymb}
\usepackage{multirow}

\usepackage{xcolor}
\usepackage[ruled,vlined,boxed]{algorithm2e}

\usepackage[accsupp]{axessibility}  

\definecolor{commentcolor}{RGB}{110,154,155}   

\def\eg{\emph{e.g}.}

\newcommand{\modelname}[0]{MuvieNeRF\xspace}

\newcommand{\smallsec}[1]{\vspace{0.2em}\noindent\textbf{#1}}

\usepackage[pagebackref=true,breaklinks=true,letterpaper=true,colorlinks,bookmarks=false]{hyperref}

\iccvfinalcopy 

\def\iccvPaperID{2214} 

\ificcvfinal\fi

\begin{document}

\title{Multi-task View Synthesis with Neural Radiance Fields}

\author{
{Shuhong Zheng$^{1,}$\thanks{Equal contribution}  \quad Zhipeng Bao$^{2,}$\footnotemark[1] \quad Martial Hebert$^{2}$ \quad Yu-Xiong Wang$^{1}$} \\
{ $^1$University of Illinois Urbana-Champaign \qquad $^2$Carnegie Mellon University } \\
  \texttt{\small  \{szheng36, yxw\}@illinois.edu \qquad \{zbao, hebert\}@cs.cmu.edu } \\
}

\maketitle
\ificcvfinal\thispagestyle{empty}\fi

\begin{abstract}
     Multi-task visual learning is a critical aspect of computer vision. Current research, however, predominantly concentrates on the multi-task dense prediction setting, which overlooks the intrinsic 3D world and its multi-view consistent structures, and lacks the capability for versatile imagination.
     In response to these limitations, we present a novel problem setting -- multi-task view synthesis (MTVS), which reinterprets multi-task prediction as a set of novel-view synthesis tasks for multiple scene properties, including RGB.
     To tackle the MTVS problem, we propose \modelname, a framework that incorporates both multi-task and cross-view knowledge to simultaneously synthesize multiple scene properties. \modelname integrates two key modules, the Cross-Task Attention (CTA) and Cross-View Attention (CVA) modules, enabling the efficient use of information across multiple views and tasks.
     Extensive evaluation on both synthetic and realistic benchmarks demonstrates that \modelname is capable of simultaneously synthesizing different scene properties with promising visual quality, even outperforming conventional discriminative models in various settings. Notably, we show that \modelname exhibits universal applicability across a range of NeRF backbones.
     Our code is available at \url{https://github.com/zsh2000/MuvieNeRF}.
\end{abstract}

\section{Introduction}
\label{sec:intro}

\begin{figure}[t]
	\centering
        \includegraphics[width =\linewidth]{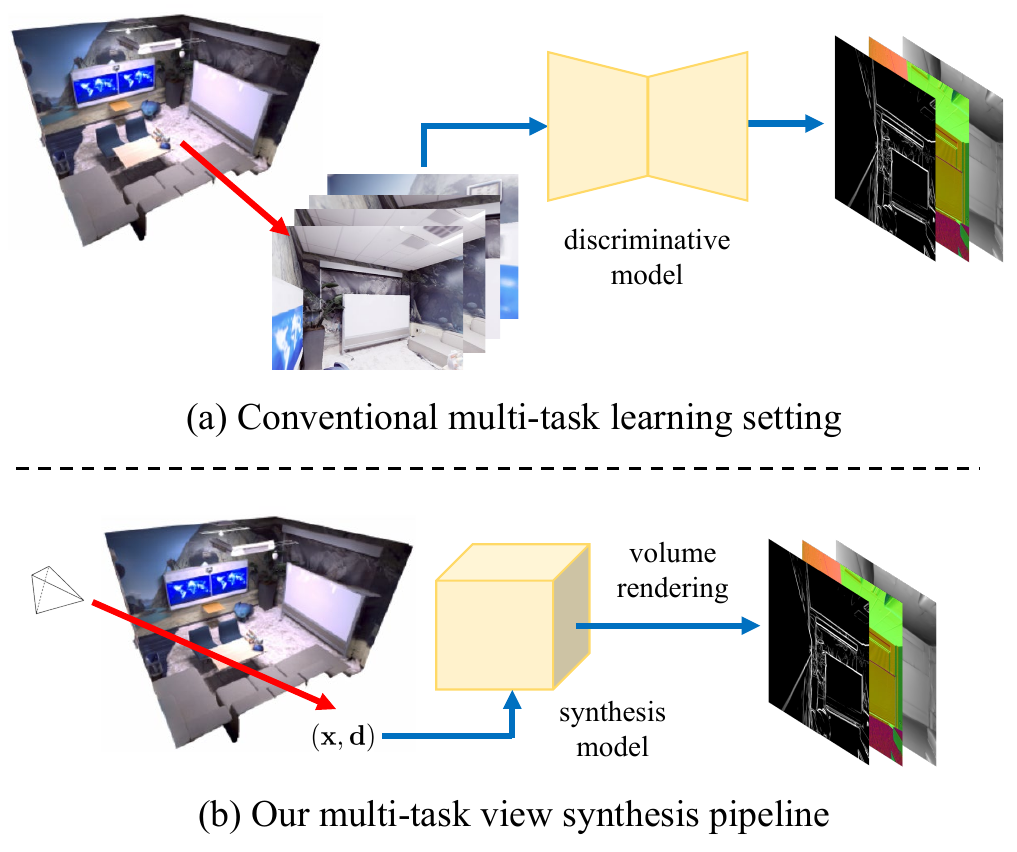}
        \vspace{-5pt}
		\caption{Comparison between (a) the conventional multi-task learning scheme and (b) our multi-task view synthesis setting. The conventional ``discriminative" multi-task learning makes predictions for single images while multi-task view synthesis aims to render visualizations for multiple scene properties at novel views.}
		\label{fig:teaser}
\end{figure}

When observing a given scene, human minds exhibit a remarkable capability to mentally simulate the objects within it from a novel viewpoint in a {\em versatile} manner~\cite{pelaprat2011minding}. It not only includes imagination of the colors of objects, but also extends to numerous associated scene properties such as surface orientation, semantic segmentation, and edge patterns. Prompted by this, a burgeoning interest has emerged, seeking to equip modern robotic systems with similar capabilities for handling multiple tasks. Nevertheless, contemporary research~\cite{misra2016cross, zamir2020robust, Zamir_2018_CVPR} has primarily centered on the {\em multi-task dense prediction} setting, which employs a conventional discriminative model to simultaneously predict multiple pixel-level scene properties using given RGB images (refer to Figure~\ref{fig:teaser}(a)). Yet, the methodologies arising from this context often demonstrate practical limitations, primarily due to their tendency to treat each image as a separate entity, without constructing an explicit 3D model that adheres to the principle of multi-view consistency. Even more critically, they lack the ability to ``imagine'' -- they are incapable of inferring scene properties from an \emph{unseen} viewpoint, as these models invariably require RGB images.

To circumvent these constraints, we propose a novel approach that revisits multi-task learning (MTL)~\cite{1997multitask} from a {\em synthesis} perspective. This leads to a more flexible problem setting that reinterprets multi-task visual learning as a collection of novel-view synthesis problems, which we refer to as {\em multi-task view synthesis} (MTVS) (refer to Figure~\ref{fig:teaser}(b)). As an illustration, the task of predicting surface normals for a given image could be reframed as visualizing a three-channel ``image'' with the given pose and camera parameters. With the achievements of Neural Radiance Fields (NeRF)~\cite{mildenhall2020nerf}, the implicit scene representation offers an effective solution to synthesize scene properties beyond RGB~\cite{ssnerf2023}. Importantly, this scene representation takes multi-view geometry into account, which consequently enhances the performance of all tasks.

Diverging from~\cite{ssnerf2023}, we make the exploration that mining multi-task knowledge can simultaneously enhance the learning of different tasks, extending beyond discriminative models~\cite{standley2020tasks, Zamir_2018_CVPR} to include synthesis models as well. Furthermore, we argue that the alignment of features across multiple reference views and the target view can reinforce cross-view consistency, thereby bolstering the implicit scene representation learning. Informed by this insight, we propose \modelname, a unified framework for the MTVS task, which incorporates \textbf{\emph{Mu}}lti-task and cross-\textbf{\emph{vie}}w knowledge, thus enabling the simultaneous synthesis of multiple scene properties through a shared implicit scene representation. \modelname can be applied to an arbitrary conditional NeRF architecture and features a unified decoder with two key modules: \emph{Cross-Task Attention (CTA) module}, which investigates relationships among different scene properties, and \emph{Cross-View Attention (CVA) module}, which aligns features across multiple views. The integration of these two modules within \modelname facilitates the efficient utilization of information from multiple views and tasks, leading to better performance across all tasks.

To demonstrate the effectiveness of our approach, we first instantiate our \modelname with GeoNeRF~\cite{geonerf2022}, a state-of-the-art conditional NeRF model, and conduct comprehensive evaluations on both synthetic and real-world benchmarks. The results illustrate that \modelname is capable of solving multi-task learning in a synthesis manner, even outperforming several competitive discriminative models in different settings. Moreover, we ablate the choice of conditional NeRF backbones to illustrate the broad applicability of our framework. We further validate the individual contributions of the CVA and CTA modules by building and comparing different variants of \modelname. Finally, we demonstrate the broader applications and analysis of \modelname, such as generalization on out-of-distribution datasets.

In summary, \textbf{our contributions} are three-fold: \textbf{(1)} We pioneer a novel problem definition, multi-task view synthesis (MTVS), which reconsiders multi-task visual learning as a set of view synthesis tasks. The introduction of MTVS paves the way for robots to emulate human-like mental simulation capabilities by utilizing the implicit scene representation offered by Neural Radiance Fields (NeRF). \textbf{(2)} We present \modelname, a unified framework that employs Cross-Task Attention (CTA) and Cross-View Attention (CVA) modules to leverage cross-view and cross-task information for the MTVS problem. \textbf{(3)} Comprehensive experimental evaluations demonstrate that \modelname shows promising results for MTVS, and greatly outperforms conventional discriminative models across diverse settings.

\section{Related Work}
\label{sec:related}

In this work, we propose the \modelname model which leverages both \emph{multi-task} and \emph{cross-view} information for \emph{multi-task view synthesis}. We review the most relevant work in the areas below.

\smallsec{View Synthesis} aims to generate a target image with an arbitrary camera pose by referring to source images~\cite{tucker2020single}.  Numerous existing methods have delivered promising results in this area~\cite{bao2021bowtie,nguyen2019hologan,sitzmann2019deepvoxels,wiles2020synsin,yin2018geonet}. 
However, unlike these conventional approaches, MTVS endeavors to synthesize multiple scene properties, including RGB, from novel viewpoints.
In pursuit of a similar goal, another group of methods seeks to render multiple annotations for novel views, following a \emph{first-reconstruct-then-render} strategy~\cite{eftekhar2021omnidata,goesele2007multi,jaritz2019multi,kong2023vmap}. 
These methods typically collect or construct a 3D scene representation (\eg, mesh or point cloud) and subsequently render multiple scene properties using 3D-to-2D projection. In contrast, our work constructs an \emph{implicit} 3D scene representation using a NeRF-style model based on 2D data. This approach is more computationally efficient and, importantly, our implicit representation provides an opportunity to further model task relationships, an advantage the aforementioned methods do not possess.

\smallsec{Neural Radiance Fields} are originally designed for synthesizing novel-view images with ray tracing and volume rendering technologies~\cite{mildenhall2020nerf}. Follow-up work~\cite{barron2021mipnerf, deng2021depth,gu2021stylenerf,Hu_2022_CVPR_efficient,kurz-adanerf2022,martinbrualla2020nerfw,Niemeyer2022Regnerf,niemeyer2021giraffe,Ost_2021_CVPR,park2021nerfies,reiser2021kilonerf,tancik2022blocknerf,wei2021nerfingmvs,xiangli2022bungeenerf} further improves the image quality, optimization, and compositionality. In addition, several approaches~\cite{mvsnerf, geonerf2022,wang2022attention, yu2021pixelnerf}, namely conditional NeRFs, encode the scene information to enable the conditional generalization to novel scenes, which are more aligned with our setting. Our \modelname takes the encoders from these conditional NeRFs as backbones.
Some work has also paid attention to synthesizing other properties of scenes~\cite{kangle2023pix2pix3d,oechsle2021unisurf, verbin2022refnerf,yariv2021volume,ssnerf2023,semantic_nerf}. Among them, Semantic-NeRF~\cite{semantic_nerf} extends NeRF from synthesizing RGB images to additionally synthesizing semantic labels. SS-NeRF~\cite{ssnerf2023} further generalizes the NeRF architecture to simultaneously render RGB and different scene properties with a shared scene representation. Panoptic 3D volumetric representation~\cite{siddiqui2022panoptic} is introduced to jointly synthesize RGB and panoptic segmentation for in-the-wild images. Different from them, we tackle the novel MTVS task and leverage both {\em cross-view} and {\em cross-task} information.

\smallsec{Multi-task Learning} aims to leverage shared knowledge across different tasks to achieve optimal performance on all the tasks. 
Recent work improves multi-task learning performance by focusing on better optimization strategies~\cite{bao2022generative,2018icml_gradnorm, 2020picksign, 2020branch,opti2022icml, javaloy2022rotograd, liu2021conflict} and exploring more efficient multi-task architectures~\cite{muit_2022,kanakis2020reparameterizing, iccv21_task-switching, sharing_multitask2022}. 

\smallsec{Cross-task Relationship} is an interesting topic in multi-task learning, which aims to explore the underlying task relationships among different visual tasks~\cite{2011whomtoshare}. Taking the task relationship into consideration, cross-stitch networks~\cite{misra2016cross} adopt a learnable parameter-sharing strategy for multi-task learning. 
Taskonomy and its follow-up work~\cite{standley2020tasks,zamir2020robust,Zamir_2018_CVPR} systematically study the internal task relationships and design the optimal multi-task learning schemes accordingly to obtain the best performance. Inspired by them, we also investigate how to better model multi-task learning but in a \textit{synthesis} framework with our model-agnostic \modelname.

\section{Method}
\label{sec:method}
In this section, we first describe our novel multi-task view synthesis problem in Section~\ref{sec:problem}. Next, we briefly review conditional neural radiance fields (NeRFs) and volume rendering in Section~\ref{sec:nerf}. In Section~\ref{sec:coconerf}, we explain the proposed \modelname (as shown in Figure~\ref{fig:model_arch}) in detail. Finally, we discuss how we handle a more challenging setting without access to source-view annotations at test time in Section~\ref{sec:challenge}.

\subsection{Multi-task View Synthesis Problem}
\label{sec:problem}

Different from conventional multi-task learning settings, our goal is to jointly synthesize multiple scene properties including RGB images from \emph{novel} views. Therefore, we aim to learn a model $\Phi$ which takes a set of $V$ source-view task annotations with camera poses as reference, and predicts the task annotations for a novel view given camera pose \textbf{(Inference Setting I)}:
\begin{equation}
    \mathbf{Y}_T = \Phi \left( \left \{(\mathbf{Y}_{i}, \mathbf{P}_{i})\right \}_{i=1}^V , \mathbf{P}_T\right),
    \label{eq:problem}
\end{equation}
where $\mathbf{Y}_i = \left[ \mathbf{x}_i, \mathbf{y}^1_i, \cdots, \mathbf{y}^K_i \right]$ denotes RGB images $\mathbf{x}_i$ and $K$ other multi-task annotations $\{\mathbf{y}^j_i\}_{j=1}^K$ in the $i^{\rm th}$ source view. $\mathbf{P}_i$ is the $i^{\rm th}$ source camera pose, and $\mathbf{P}_T$ is the target camera pose. 
During the evaluation, $\Phi$ is \textit{supposed to be generalized to novel scenes that are not seen during training}.

For the evaluation of those novel scenes, we also provide a more challenging setting lacking source-view annotations during the inference time with the assumption that the model may not get access to additional annotations other than RGB during inference \textbf{(Inference Setting II)}:
\begin{equation}
    \mathbf{Y}_T = \Phi \left( \left \{(\mathbf{x}_{i}, \mathbf{P}_{i})\right \}_{i=1}^V , \mathbf{P}_T\right).
    \label{eq:challenging}
\end{equation}

With the above two settings, \textbf{Inference Setting I} allows us to better evaluate the task relationships in our synthesis framework in a cleaner manner, so it is \textit{the focused setting} in our paper; \textbf{Inference Setting II} is more aligned with real practice, for which we also propose a solution which is discussed in Sections~\ref{sec:challenge} and~\ref{sec:exploration}.

\subsection{Preliminary: Conditional Neural Radiance Fields and Volume Rendering}
\label{sec:nerf}

\begin{figure*}[t]
		\centering
        \includegraphics[width =0.95\linewidth]{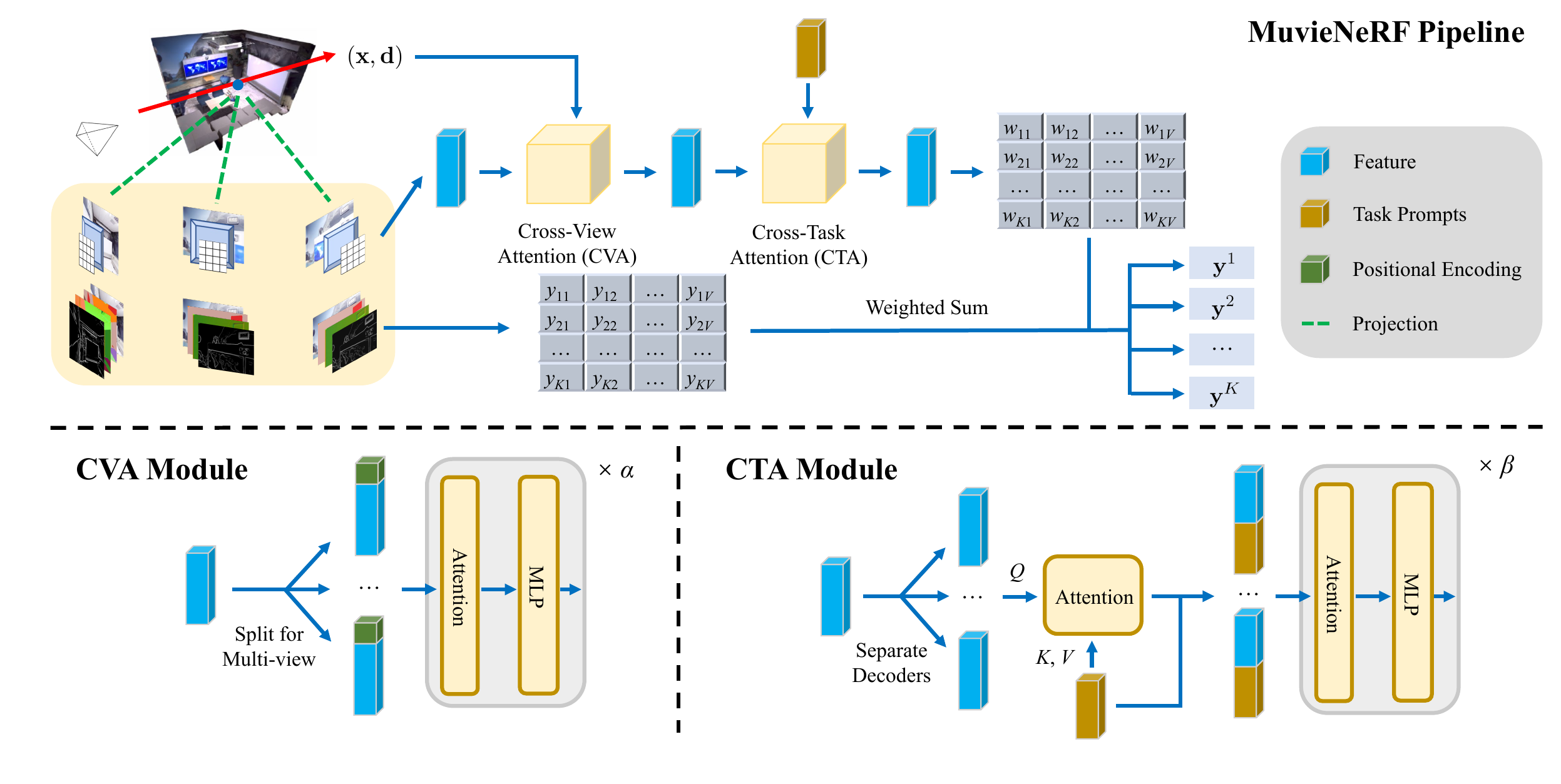}	
          \vspace{-10pt}
		\caption{Model architecture. \modelname is a unified framework for multi-task view synthesis equipped with Cross-View Attention (CVA) and Cross-Task Attention (CTA) modules. It predicts multiple scene properties for arbitrary 3D coordinates with source-view annotations.
  }
  \vspace{-10pt}
\label{fig:model_arch}
\end{figure*}

\textbf{Neural radiance fields (NeRFs)}~\cite{mildenhall2020nerf} propose a powerful solution for implicit scene representation, and are widely used in novel view image synthesis. Given the 3D position of a point $\mathbf{q} = (x, y, z)$ in the scene and the 2D viewing direction $\mathbf d = (\theta, \phi)$, NeRFs learn a mapping function $(\mathbf{c}, \sigma) = F(\mathbf{q}, \mathbf{d})$, which maps the 5D input $(\mathbf{q}, \mathbf{d})$ to RGB color $\mathbf c=(r, g, b)$ and density $\sigma$. 

To enhance the generalizability of NeRFs, \textbf{conditional NeRFs}~\cite{mvsnerf,geonerf2022,wang2022attention,yu2021pixelnerf} learn a scene representation across multiple scenes. They first extract a feature volume $\mathbf{W} = E(\mathbf{\mathbf{x}})$ for each input image $\mathbf{x}$ of a scene. Next, for an arbitrary point $\mathbf{q}$ on a camera ray with direction $\mathbf{d}$, they are able to retrieve the corresponding image feature on $\mathbf{W}$ by projecting $\mathbf{q}$ onto the image plane with known pose $\mathbf{P}$. We treat the above part as the \emph{conditional NeRF encoder}, which returns:
\begin{equation}
    f_{\rm scene} = F_{\rm enc}(\left\{ \mathbf{x}_i, \mathbf{P}_i\right\}_{i=1}^V, \mathbf{q}).
\end{equation}
We have $f_{\rm scene} \in \mathbb{R}^{V \times d_\mathrm{scene}}$, which contains the scene representation from $V$ views. Next, the conditional NeRFs further learn a decoder $(\mathbf{c}, \sigma) = F_{\rm dec}(\mathbf{q}, \mathbf{d}, f_{\rm scene})$ to predict the color and density. 

Given the color and density of 3D points, NeRFs render the 2D images by running \textbf{volume rendering} for each pixel with ray tracing. Every time when rendering a pixel in a certain view, a ray $\mathbf{r}(t) = \mathbf{o} + t\mathbf{d}$ which origins from the center $\mathbf{o}$ of the camera plane in the direction $\mathbf{d}$ is traced. NeRFs randomly sample $M$ points $\{t_m \}_{m=1}^M$ with color $\mathbf c(t_m)$ and density $\sigma(t_m)$ between the near boundary $t_n$ and far boundary $t_f$. The RGB value of the pixel is given by:
\begin{equation}
    \mathbf{\hat{C}}(\mathbf{r}) = \sum_{m=1}^M \hat{T}(t_m)\alpha (\delta_m\sigma(t_m))\mathbf{c}(t_m),
\end{equation}
where $\delta_m$ is the distance between two consecutive sampled points ($\delta_m =\| t_{m+1} - t_m \|$), $\alpha(d) = 1 - \exp(-d)$, and 
\begin{equation}
    \hat{T}(t_m) = \exp\left(- \sum_{j = 1}^{m-1} \delta_j \sigma(t_j)  \right)
\end{equation}
denotes the accumulated transmittance. The same technique can be used to render an arbitrary scene property $\mathbf{y}^j$ by:
\begin{equation}
    \mathbf{\hat{Y}}^j(\mathbf{r}) = \sum_{m=1}^M \hat{T}(t_m)\alpha (\delta_m\sigma(t_m))\mathbf{y}^j(t_m).
\label{eq:render}
\end{equation}

\subsection{\modelname}
\label{sec:coconerf}
As illustrated in Figure~\ref{fig:model_arch}, \modelname first fetches the scene representation $f_{\rm scene}$ from the conditional NeRF encoder, then predicts multiple scene properties $\left [ \mathbf{x}(\mathbf{q}), \mathbf{y}^1(\mathbf{q}), \cdots, \mathbf{y}^K(\mathbf{q}) \right]$ for arbitrary 3D coordinate $\mathbf{q}$. The final annotations are rendered by Equation~\ref{eq:render}. We explain how to predict multiple scene properties with $f_{\rm scene}$ and source annotations $[(\mathbf{Y}_1,\mathbf{P}_1), \cdots, (\mathbf{Y}_V,\mathbf{P}_V)] $ as follows. The full detailed architecture is included in the appendix.

\subsubsection{Cross-View Attention Module}

The cross-view attention (CVA) module (Figure~\ref{fig:model_arch} bottom left) leverages the multi-view information for \modelname. To start, we first concatenate $f_{\rm scene}$ with a positional embedding derived from the target ray and the source-view image plane: $f_{\rm scene}^{\rm pos} = \left[ f_{\rm scene}; \gamma(\theta_{n,v}) \right]$, where $\gamma(\cdot)$ is the positional encoding proposed
in~\cite{mildenhall2020nerf}, and $\theta_{n,v}$ is the angle between the novel camera ray $\mathbf{r}$ and the line that connects the camera center of view $v$ and the point $\mathbf{q}_{n}$ in the queried ray, which measures the similarity between the source view $v$ and the target view.

Next, $\alpha$ CVA modules are used to leverage the cross-view information. Concretely, in each module, we have one self-attention union followed by a multi-layer perceptron (MLP): $f_{\rm CVA} = \mathrm{MLP}_\mathrm{CVA}(f_{\rm scene}^{\rm pos} + \mathrm{MHA}(f_{\rm scene}^{\rm pos},f_{\rm scene}^{\rm pos}))$,
where $\mathrm{MHA}(a,b)$ denotes multi-head attention~\cite{attention17} with $a$ as query and $b$ as key and value. 

After these processes, we apply $(K+1)$ different MLPs (corresponding to $K$ vision tasks in multi-task view synthesis plus the RGB synthesis task) to broadcast the shared feature, leading to the $(K+1)$-branch feature $f_{\rm task} \in \mathbb{R}^{(K+1) \times V \times d_\mathrm{task}}$.

\begin{table*}[t]
\centering
    \resizebox{\linewidth}{!}{
    \begin{tabular}{l|l|cccccc|cccccc}
     \multicolumn{2}{c|}{Evaluation Type} & \multicolumn{6}{c|}{Training scene evaluation} & \multicolumn{6}{c}{Testing scene evaluation} \\     \hline 
      \multicolumn{2}{c|}{Task} & RGB ($\uparrow$)                    & SN ($\downarrow$) & SH ($\downarrow$) & ED ($\downarrow$) & KP ($\downarrow$) & SL ($\uparrow$) & RGB ($\uparrow$)                    & SN ($\downarrow$) & SH ($\downarrow$) & ED ($\downarrow$) & KP ($\downarrow$) & SL ($\uparrow$)\\ \hline
     \multirow{4}{*}{Replica} & Heuristic &  29.60 & 0.0272  & 0.0482 & 0.0214 & 0.0049 & 0.9325 & 20.86 & 0.0395 & 0.0515 & 0.0471 & 0.0097 & 0.8543 \\ 
     & Semantic-NeRF & 33.60 & 0.0211 & 0.0403 & 0.0128 & 0.0037 & 0.9507 & 27.08 & 0.0221 & 0.0418 & 0.0212 & 0.0055 & 0.9417\\
     & SS-NeRF & 33.76 & 0.0212 & 0.0383 & 0.0116 & 0.0035 & 0.9533 & 27.22 & 0.0224 & \textbf{0.0405} & 0.0196 & 0.0053 & 0.9483 \\
     & \modelname & \textbf{34.92} & \textbf{0.0193} & \textbf{0.0345} & \textbf{0.0100} & \textbf{0.0034} & \textbf{0.9582} & \textbf{28.55} & \textbf{0.0201} & 0.0408 & \textbf{0.0162} & \textbf{0.0051} & \textbf{0.9563}\\ \hline 
     \multirow{4}{*}{\shortstack{ SceneNet \\ RGB-D}} & Heuristic & 22.66 & 0.0496 & - & 0.0521 & 0.0093 & 0.8687 & 22.02 & 0.0394 & - & 0.0525 & 0.0124 & 0.8917 \\ 
     & Semantic-NeRF & 28.29 & 0.0248 & - & 0.0212 & 0.0050 & 0.9152 & 28.85 & 0.0186 & - & 0.0198 & 0.0051 & 0.9417 \\
     & SS-NeRF & 28.93 & 0.0244& - & 0.0216 & 0.0050 & 0.9175 & 29.18 & 0.0182 & - & 0.0197 & 0.0052 & 0.9510 \\
     & \modelname &  \textbf{29.29} & \textbf{0.0237} & - & \textbf{0.0207} & \textbf{0.0049} & \textbf{0.9190} & \textbf{29.56} & \textbf{0.0173} & - & \textbf{0.0189} & \textbf{0.0050} & \textbf{0.9556} \\
\end{tabular}
}
\vspace{2pt}
    \caption{Averaged performance of \modelname on Replica~\cite{straub2019replica} and SceneNet RGB-D~\cite{mccormac2016scenenet} datasets on both training scenes and testing scenes. Full results with multiple runs are provided in the appendix, our model consistently outperforms both the single-task Semantic-NeRF baseline and multi-task SS-NeRF baseline, owing to the proposed CVA and CTA modules.}
    \label{tab:main_result}
\end{table*}

\subsubsection{Cross-Task Attention Module}
In order to simultaneously benefit all the downstream tasks, we propose a novel cross-task attention (CTA) module (Figure~\ref{fig:model_arch} bottom right) to facilitate knowledge sharing and information flow among all the tasks. The CTA module has two attention components with shared learnable task prompts~\cite{ye2023taskprompter}, $p_t\in \mathbb{R}^{(K+1) \times d_{t}}$, where $d_t$ is the dimension of task prompts. The first attention component applies cross-attention between features from each branch and the task prompts $f_\mathrm{stage1} = f_{\rm task} + \mathrm{MHA}(f_{\rm task},p_t)$. In this stage, we run $K$ MHA individually for each task branch with the shared task prompts. After the cross-attention, we further concatenate $f_\mathrm{stage1}^j$ for task $T_j$ and the corresponding task prompt $p_t^j$ to obtain $f_\mathrm{stage1'}$.

Next, we apply the second component to use $\beta$ self-attention modules for all the branches jointly to leverage the cross-task features. The final feature representation is obtained by: $f_\mathrm{stage2} = \mathrm{MLP}_\mathrm{CTA}(f_\mathrm{stage1'} + \mathrm{MHA}(f_\mathrm{stage1'},f_\mathrm{stage1'}))$.

Finally, to predict the task annotations of the target view, we adopt the formulation of GeoNeRF~\cite{geonerf2022}. The prediction $\hat{\mathbf{y}}^{j}$ of task $T_{j}$ on the target view is the weighted sum of the source views:
\begin{equation}
    \hat{\mathbf{y}}^{j} = \sum_{i=1}^V \mathbf{w}[j,i] \cdot \mathbf{y}[j,i],
\end{equation}
where the matrix $\mathbf{y}$ is made of input view annotations $\left\{\mathbf{Y}_{i}\right\}_{i=1}^{V}$ and $\mathbf{w}$ is obtained by an additional MLP layer which processes $f_\mathrm{stage2}$. 

\subsubsection{Optimization}

For the set of $K$ tasks $\mathcal{T} = \{ T_1, T_2, \cdots, T_K \}$ including the RGB colors, we apply their objectives individually and the final objective is formulated as 
\begin{equation}
\mathcal{L}_\mathrm{MT} = \sum_{T_j \in \mathcal{T}} \lambda_{T_j} \mathcal{L}_{T_j},
\end{equation}
where $\lambda_{T_j}$ is the weight for the corresponding task $T_j$. For each task, $\mathcal{L}_{T_j}$ is formulated as:
\begin{equation}
\mathcal{L}_{T_j}=\sum_{\mathbf{r}\in \mathcal{R}} \mathcal{L}_j(\hat{\mathbf{y}}^j(\mathbf{r}), \mathbf{y}^{j}(\mathbf{r})),
\end{equation}
where $\mathbf{y}^j(\mathbf{r}), \hat{\mathbf{y}}^j(\mathbf{r})$ are the ground-truth and prediction for a single pixel regarding task $T_j$. $\mathcal{R}$ is the set of rays $\mathbf{r}$ for all training views. $\mathcal{L}_j$ is chosen from $L_1$ loss, $L_2$ loss, and cross-entropy loss according to the characteristics of the tasks.

\subsection{Tackling without Source-view Annotations}
\label{sec:challenge}

The proposed model is based on the assumption that source-view task annotations are available during \emph{inference} time. The assumption rules out the influence of inaccurate source-view task information, which sets a cleaner environment to excavate multi-task synergy in a synthesis framework for the MTVS problem. However, from the real application perspective, traditional discriminative models only take RGB images as input without any task annotations. To demonstrate that our model is able to be applied in real scenarios, we introduce the more challenging \textbf{Inference Setting II} formulated by Equation~\ref{eq:challenging} and provide a solution by incorporating a U-Net~\cite{ronneberger2015u} shaped module $F_\mathrm{UNet}$ into our \modelname architecture. The detailed architecture of $F_\mathrm{UNet}$ is shown in the appendix.

Conceptually, $F_\mathrm{UNet}$ takes RGB images from the $V$ source views $\{\mathbf{x}_i\}_{i=1}^V$ as input and produces the corresponding multi-task annotations $\{\Tilde{\mathbf{Y}}_i\}_{i=1}^V$, where $\Tilde{\mathbf{Y}}_i = [\Tilde{\mathbf{y}}_i^1, \cdots, \Tilde{\mathbf{y}}_i^K]$. Next, similar to the conditional NeRF encoder, we retrieve the corresponding multi-task annotations $\{\Tilde{\mathbf{Y}}_i(\mathbf{q})\}_{i=1}^V$ for an arbitrary point $\mathbf{q}$ by projection. 

During training time, $F_\mathrm{UNet}$ is trained with pixel-wise task-specific losses. Concretely, for task $T_j$, we have:
\begin{equation}
\mathcal{L}_{U_j}=\sum_{\mathbf{r}\in \mathcal{R}} \sum_{i=1}^V \mathcal{L}_j(\Tilde{\mathbf{y}}_i^j(\mathbf{r}), \mathbf{y}_i^{j}(\mathbf{r})).
\end{equation}
The final loss becomes $\mathcal{L}_\mathrm{final} = \sum_{T_j \in \mathcal{T}} \lambda_{T_j} (\mathcal{L}_{T_j} + \mathcal{L}_{U_j})$, for which we take the ground-truth multi-task annotations to learn the weights during training. However, we instead use the predictions produced by $F_\mathrm{UNet}$ for inference:
\begin{equation}
    \hat{\mathbf{y}}^{j} = \sum_{i=1}^V \mathbf{w}[j,i] \cdot \Tilde{\mathbf{y}}[j,i].
    \label{eq:weightedsum}
\end{equation}

\section{Experimental Evaluation}
\label{sec:experiments}
In this section, we start with main evaluation from Sections~\ref{sec:setting} to \ref{sec:discriminative}, including experimental setting, quantitative and qualitative results, and comparison with conventional discriminative multi-task models. Next, we make further investigations in Sections~\ref{sec:ablation} and \ref{sec:exploration}, including ablation studies and additional explorations. Finally, we discuss the limitations and future work in Section~\ref{sec:limitation}.

\subsection{Experimental Setting}
\label{sec:setting}

\smallsec{Model Instantiation:} As illustrated in Section~\ref{sec:method}, our model can build upon arbitrary conditional NeRF encoders. For the main evaluation, we instantiate our model with state-of-the-art GeoNeRF~\cite{geonerf2022}. We use $\alpha = 4$ and $\beta = 2$ for the number of self-attention unions in the CVA and CTA modules, respectively. We additionally show the performance with other NeRF encoders in the ablation study. 

\smallsec{Benchmarks:} We take two benchmarks for our main evaluation.  \textbf{Replica}~\cite{straub2019replica} is a commonly-used indoor scene dataset containing high-quality photo-realistic 3D modelling of 18 scenes. Following the data acquisition method as~\cite{semantic_nerf}, we collect 22 sequences from the scenes, each containing 50 frames at a resolution of $640\times480$. \textbf{SceneNet RGB-D}~\cite{mccormac2016scenenet} is a large-scale photorealistic indoor scene dataset expanding from SceneNet~\cite{Handa:etal:arXiv2016}. We include 32 scenes with 40 frames of each at a resolution of $320 \times 240$ in our evaluation. In addition to the above two datasets, we further evaluate \textit{zero-shot} adaptation on four \textit{out-of-distribution} datasets: LLFF~\cite{mildenhall2019local}, TartanAir~\cite{tartanair}, ScanNet~\cite{dai2017scannet}, and BlendedMVS~\cite{yao2020blendedmvs}.

\begin{figure*}[t]
		\centering
        \includegraphics[width =0.9\linewidth]{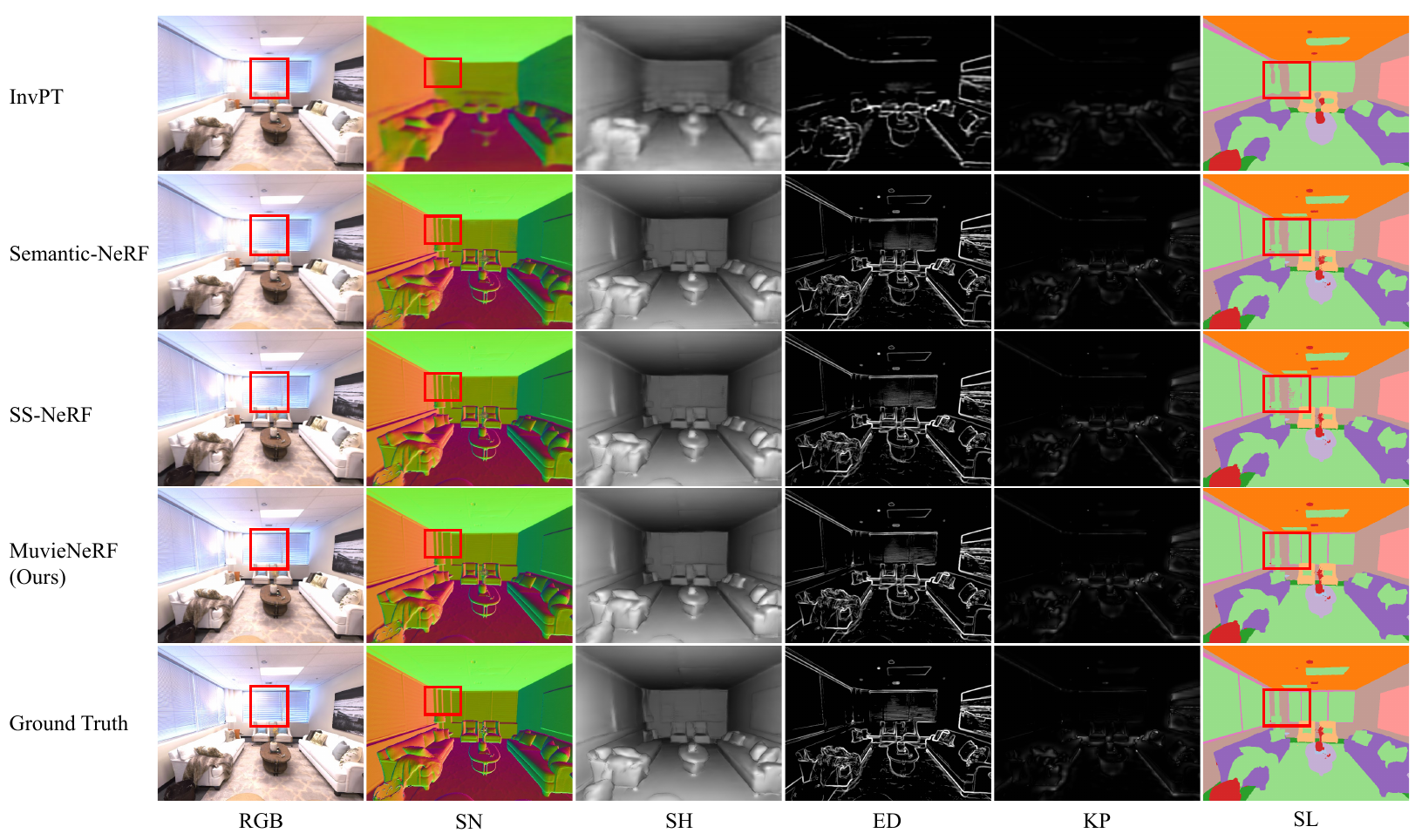}	
        \vspace{-2 pt}
		\caption{Visual comparisons of our model and baselines on a test scene of Replica dataset~\cite{straub2019replica}. Our predictions are sharper and more accurate compared with other baselines. The underlying reason is that shared cross-view and cross-task knowledge can provide additional information for the target tasks.
  }
		\label{fig:main_result}
\end{figure*}

\begin{table}[t]
\centering
    \resizebox{\linewidth}{!}{
    \begin{tabular}{l|cccccc}
      Model & RGB ($\uparrow$)                    & SN ($\downarrow$) & SH ($\downarrow$) & ED ($\downarrow$) & KP ($\downarrow$) & SL ($\uparrow$) \\ \hline
      $\text{\modelname}_\mathrm{w/o~SH}$ & 28.26 & 0.0204 & - & 0.0171 & \textbf{0.0051} & 0.9557 \\
      $\text{\modelname}_\mathrm{w/o~KP}$ & 27.96 & 0.0212 & 0.0423 & 0.0181 & - & 0.9519 \\ \hline
     \modelname  & \textbf{28.55} & \textbf{0.0201} & \textbf{0.0408} & \textbf{0.0162} & \textbf{0.0051} & \textbf{0.9563} \\ 
\end{tabular}
}
\vspace{5pt}
    \caption{Additional test scene evaluation for our variants without SH ($\text{\modelname}_\mathrm{w/o~SH}$) and KP ($\text{\modelname}_\mathrm{w/o~KP}$) tasks on the Replica dataset. 
    These two tasks work as a role of auxiliary tasks.}
    \label{tab:w/oKPSH}
\end{table}

\smallsec{Task Selection:} We select six representative tasks to evaluate our method following previous multi-task learning pipelines~\cite{standley2020tasks,ssnerf2023}. The tasks are Surface Normal Prediction \textbf{(SN)}, Shading Estimation \textbf{(SH)}, Edges Detection \textbf{(ED)}, Keypoints Detection \textbf{(KP)}, and Semantic Labeling \textbf{(SL)}, together with \textbf{RGB} synthesis. For the SceneNet RGB-D dataset, we drop the SH task due to missing annotations.

\smallsec{Evaluation Setup:} For the Replica dataset, we divide the 22 scenes into 18, 1, and 3 for training, validation, and testing, respectively. For SceneNet RGB-D, we split 26 scenes for training, 2 for validation, and 4 for testing. For each scene, we hold out every 8 frames as testing views.

For these held-out views, we provide two types of evaluation: \textit{Training scene evaluation} is conducted on novel views from the training scenes; \textit{Testing scene evaluation} runs on novel scenes and is used to evaluate the generalization capability of the compared models.

\smallsec{Evaluation Metrics:} For RGB synthesis, we measure Peak Signal-to-Noise Ratio (PSNR) for evaluation. For semantic segmentation, we take mean Intersection-over-Union (mIoU). For the other tasks, we evaluate the $L_1$ error.

\smallsec{Baselines:} We consider synthesis baselines for the main evaluation. \textbf{Semantic-NeRF}~\cite{semantic_nerf} extends NeRF for the semantic segmentation task. We further extend this model in the same way for other tasks, which only considers single-task learning in a NeRF style. \textbf{SS-NeRF}~\cite{ssnerf2023} considers multi-task learning in a NeRF style, but ignores the cross-view and cross-task information. We equip both models with the same GeoNeRF backbone as our model. Following~\cite{ssnerf2023}, we also include a \textbf{Heuristic} baseline which estimates the annotations of the test view by projecting the source labels from the nearest training view to the target view. 

\smallsec{Implementation Details:} 
We set the weights for the six chosen tasks as $\lambda_{{\rm RGB}}=1$, $\lambda_{{\rm SN}}=1$, $\lambda_{\rm SL}=0.04$, $\lambda_{\rm SH}=0.1$, $\lambda_{\rm KP}=2$, and $\lambda_{\rm ED}=0.4$ based on empirical observations. We use the Adam~\cite{kingma2014adam} optimizer with an initial learning rate of $5\times 10^{-4}$ and set $\beta_{1}=0.9, \beta_{2}=0.999$. During training, each iteration contains a batch size of 1,024 rays randomly sampled from all training scenes. 

More details about our encoder architectures, dataset processing, out-of-distribution analysis, and implementation are included in the appendix.

\begin{table*}[t]
\centering
    \resizebox{\linewidth}{!}{
    \begin{tabular}{l|ccccc|ccccc|ccccc}
     \multirow{2}{*}{Model} & \multicolumn{5}{c|}{NeRF's Images (No Tuned)} & \multicolumn{5}{c|}{NeRF's Images (Tuned)} & \multicolumn{5}{c}{GT Images (Upper Bound)} \\     \cline{2-16} 
      & SN ($\downarrow$) & SH ($\downarrow$) & ED ($\downarrow$) & KP ($\downarrow$) & SL ($\uparrow$)                   & SN ($\downarrow$) & SH ($\downarrow$) & ED ($\downarrow$) & KP ($\downarrow$) & SL ($\uparrow$) & SN ($\downarrow$) & SH ($\downarrow$) & ED ($\downarrow$) & KP ($\downarrow$) & SL ($\uparrow$)\\ \hline
    Taskgrouping & 0.0568 & 0.0707 & 0.0408 & 0.0089 & 0.5361 & 0.0530 & 0.0677 & 0.0423 & 0.0090 & 0.5590 & 0.0496 & 0.0607 & 0.0298 & 0.0060 & 0.6191 \\ 
      MTI-Net & 0.0560 & 0.0636 & 0.0418 & \bf 0.0078 & 0.5440 & 0.0486 & \bf 0.0549 & 0.0389 & 0.0078 & 0.6753 & 0.0422 &  0.0498 & \bf 0.0281 & \bf 0.0050 & 0.7196 \\
      InvPT & \bf 0.0479 &\bf 0.0618 & \bf 0.0400 & 0.0091 & \bf 0.7139 & \bf 0.0474 & 0.0587 & \bf 0.0328 & \bf 0.0074 & \bf 0.7084 & \bf 0.0409 & \bf 0.0484 & 0.0282 & 0.0055 & \bf 0.8158 \\ \hline
      Ours & \bf 0.0201 & \bf 0.0408 & \bf 0.0162 & \bf 0.0051 & \bf 0.9563 & - & - & - & - & - & - & - & - & - & -
\end{tabular}
}
\vspace{1pt}
    \caption{Comparison to the discriminative models for the test scenes on Replica~\cite{straub2019replica} dataset. \modelname clearly beats all the discriminative models in all three settings, indicating that our model is more capable of both performance and generalizability.}
    \label{tab:discriminative}
\end{table*}

\subsection{\modelname Is Capable of Solving MTVS}
\label{sec:results}

In Table~\ref{tab:main_result}, we present the average results derived from the held-out views across both the training and testing scenes. The key observations are, firstly, it is clear that our problem statement is non-trivial as evidenced by the notably inferior performance exhibited by the simple heuristic baseline when compared to the other models. Secondly, SS-NeRF surpasses Semantic-NeRF on average, indicating the contribution of multi-task learning. Lastly, our model consistently outperforms all the baselines, reaffirming that cross-view and cross-task information are invariably beneficial within our framework.

Interestingly, we note that \modelname exhibits a performance closely comparable to the two NeRF baselines on novel scenes for the KP and SH tasks. To decipher the underlying reason, we carry out an additional evaluation on two variants of our model without the two tasks, $\text{\modelname}_\mathrm{w/o~KP}$ and $\text{\modelname}_\mathrm{w/o~SH}$, on the test scenes of Replica in Table~\ref{tab:w/oKPSH}. Our findings indicate that the KP and SH tasks indeed enhance the learning of other tasks, serving as effective auxiliary tasks. This conclusion aligns with previous studies on traditional multi-task learning models as reported by~\cite{standley2020tasks}.

Figure~\ref{fig:main_result} showcases a comparative analysis of the qualitative results. It is evident that our predictions supersede those of other baselines in terms of precision and clarity. This superior performance can be attributed to the additional information provided by shared cross-view and cross-task knowledge, which proves beneficial for the target tasks.

\begin{table}[t]
\centering
    \resizebox{\linewidth}{!}{
    \begin{tabular}{l|cccccc}
    Model & RGB ($\uparrow$) & SN ($\downarrow$) & SH ($\downarrow$) & ED ($\downarrow$) & KP ($\downarrow$) & SL ($\uparrow$) \\ \hline
    $\text{\modelname}_\mathrm{w/o~CTA}$ & 27.55 & 0.0214 & 0.0424 & 0.0198 & 0.0056 & 0.9501 \\ 
      $\text{\modelname}_\mathrm{w/o~CVA}$ & 28.25 & 0.0206  & \textbf{0.0407} & 0.0170 & 0.0052 & 0.9557\\
      \modelname & \textbf{28.55} & \textbf{0.0201} & 0.0408 & \textbf{0.0162} & \textbf{0.0051} & \textbf{0.9563} \\
\end{tabular}
}
\vspace{3pt}
    \caption{Ablation study with CTA and CVA modules on  Replica~\cite{straub2019replica} dataset. $\text{\modelname}_\mathrm{w/o~CTA}$ is the variant without CTA module; $\text{\modelname}_\mathrm{w/o~CVA}$ is the variant without CVA module. The CTA module is more crucial compared to the CVA module while combining them leads to the best performance. }
    \label{tab:ablation}
\end{table}

\begin{table}[t]
\centering
    \resizebox{\linewidth}{!}{
    \begin{tabular}{l|cccccc}
    Backbone & RGB ($\uparrow$) & SN ($\downarrow$) & SH ($\downarrow$) & ED ($\downarrow$) & KP ($\downarrow$) & SL ($\uparrow$) \\ \hline
    PixelNeRF + SS-NeRF & 23.21 & 0.0328 & 0.0457 & 0.0376 & 0.0074 & 0.8620 \\ 
    PixelNeRF + Ours & 24.14 & 0.0302 & 0.0420 & 0.0342 & 0.0068 & 0.8961\\ 
      GNT + SS-NeRF & 21.51 & 0.0405 & 0.0497 & 0.0420 & 0.0096 & 0.8302 \\
      GNT + Ours & 22.67 & 0.0333 & 0.0455 & 0.0384 & 0.0076 & 0.8686\\
      MVSNeRF + SS-NeRF & 25.27 & 0.0261 & 0.0419 & 0.0248 & 0.0061 & 0.9294\\
      MVSNeRF + Ours & 25.73 & 0.0248 & 0.0408 & 0.0227 & 0.0056 & 0.9303\\
      \hline
      GeoNeRF + SS-NeRF & 27.22 & 0.0224 & \bf 0.0405 & 0.0196 & 0.0053 & 0.9483 \\
      GeoNeRF + Ours & \bf 28.55 & \bf 0.0201 & 0.0408 & \bf 0.0162 & \bf 0.0051 & \bf 0.9563 
\end{tabular}
}
\vspace{3pt}
    \caption{Ablation study with different choices of conditional NeRF encoders. The proposed \modelname is universally beneficial to different encoders, owing to the proposed CTA and CVA modules.}
    \label{tab:nerf_encoder}
\end{table}

\begin{table}[t]
\centering
    \resizebox{0.6 \linewidth}{!}{
    \begin{tabular}{l|ccc}
    Model & SN ($\downarrow$) & ED ($\downarrow$) & KP ($\downarrow$) \\ \hline
    Discriminative & 0.0778 & 0.0355 & 0.0086\\ 
      $\text{\modelname}_D$ & \bf 0.0605 & \bf 0.0230 & \bf 0.0074\\
\end{tabular}
}
\vspace{2pt}
    \caption{Results for the setting with unknown nearby-view annotations. $\text{\modelname}_D$ still outperforms the hybrid discriminative model with a similar backbone.}
    \label{tab:challenging} 
\end{table}

\subsection{\modelname Beats Discriminative Models}
\label{sec:discriminative}

\begin{table*}[t]
\centering
    \resizebox{\linewidth}{!}{
    \begin{tabular}{c|l|cccccc|cccccc} 
\multicolumn{2}{c|}{Evaluation Type} & \multicolumn{6}{c|}{Training scene evaluation} & \multicolumn{6}{c}{Testing scene evaluation} \\     \hline 
\multicolumn{2}{c|}{Tasks}       & RGB ($\uparrow$)                    & SN ($\downarrow$) & SH ($\downarrow$) & ED ($\downarrow$) & KP ($\downarrow$) & SL ($\uparrow$) & RGB ($\uparrow$)                    & SN ($\downarrow$) & SH ($\downarrow$) & ED ($\downarrow$) & KP ($\downarrow$) & SL ($\uparrow$)\\ \hline
\multirow{3}{*}{Replica} & 
     SS-NeRF & 33.76 & 0.0212 & 0.0383 & 0.0116 & 0.0035 & 0.9533 & 27.22 & 0.0224 & \textbf{0.0405} & 0.0196 & 0.0053 & 0.9483  \\
    & MuvieNeRF & \textbf{34.92} & \textbf{0.0193} & \textbf{0.0345} & \textbf{0.0100} & \textbf{0.0034} & \textbf{0.9582} & \textbf{28.55} & \textbf{0.0201} & 0.0408 & \textbf{0.0162} & \textbf{0.0051} & \textbf{0.9563}  \\
    & SS-NeRF (Enlarged) & 34.20 & 0.0210 & 0.0388 & 0.0107 & 0.0035 & 0.9557 & 27.37 & 0.0226 & \textbf{0.0405} &  0.0186 & 0.0053 & 0.9498 \\\hline 
\multirow{3}{*}{\shortstack{ SceneNet \\ RGB-D}} & 
 SS-NeRF & 28.93 & 0.0244& - & 0.0216 & 0.0050 & 0.9175 & 29.18 & 0.0182 & - & 0.0197 & 0.0052 & 0.9510  \\
    & MuvieNeRF & \textbf{29.29} & \textbf{0.0237} & - & \textbf{0.0207} & \textbf{0.0049} & \textbf{0.9190} & \textbf{29.56} & \textbf{0.0173} & - & \textbf{0.0189} & \textbf{0.0050} & \textbf{0.9556}  \\
   &  SS-NeRF (Enlarged) & 29.03 &  0.0244 &  - & 0.0215 & \textbf{0.0049} & 0.9186 & 29.46 & 0.0182 & - &  0.0191 & \textbf{0.0050} & 0.9520 \\
\end{tabular}
}
\vspace{2pt}
\caption{Comparison between \modelname and an enlarged version of SS-NeRF. our model still significantly outperforms SS-NeRF (Enlarged), indicating that the good performance of MuvieNeRF is not simply achieved by a heavier decoder, and demonstrating the effectiveness of our module designs.}
\vspace{-5pt}
\label{tab:enlarge}
\end{table*}

\begin{table}[t]
\centering
    \resizebox{0.95\linewidth}{!}{
    \begin{tabular}{l|ccccc}
     Model  & Runtime (Per Training Iter.)  & Num. of Params. & FLOPs   \\ \hline
    SS-NeRF & 1× &  1.21M & $4.57\times10^{11}$    \\
    MuvieNeRF & 1.22× & 1.30M & $5.84\times10^{11}$  \\
\end{tabular}
}
\vspace{2pt}
\caption{Computational cost for SS-NeRF and MuvieNeRF on Replica dataset. Our CTA and CVA modules are light-weight designs.}
\vspace{-4pt}
\label{tab:cost}
\end{table}

Although conventional discriminative models fall short in addressing the proposed MTVS problem, we have explored several hybrid settings to facilitate a comparison between our \modelname and these discriminative models.

\smallsec{Hybrid Setup:} We provide \emph{additional} RGB images from novel views to the discriminative models under three settings with different choices of RGB images. (1) We train on ground truth (GT) pairs and evaluate on novel view RGB images generated by a NeRF (\emph{NeRF's Images (No Tuned)}); (2) We additionally fine-tune the discriminative models with paired NeRF's images and corresponding GT (\emph{NeRF's Images (Tuned)}); (3) We evaluate on GT images from novel views as the performance upper bound (\emph{GT Images (Upper Bound)}). For all the settings, we train the discriminative models on both training and testing scenes (training views only) to make sure that they get access to the same number of data as \modelname.

\smallsec{Discriminative Baselines:} We select three representative baselines of different architectures. \textbf{Taskgrouping}~\cite{standley2020tasks} leverages an encoder-decoder architecture with a shared representation. \textbf{MTI-Net}~\cite{MTINet} adopts a convolutional neural network backbone and enables multi-scale task interactions. \textbf{InvPT}~\cite{invpt} takes a transformer-based architecture that encourages long-range and global context to benefit multi-task learning.

Table~\ref{tab:discriminative} details the averaged results and Figure~\ref{fig:main_result} offers a visual comparison. It is discernible that our \modelname outstrips all the discriminative models, illustrating that these models struggle to effectively tackle the MTVS problem, despite fine-tuning or utilizing GT images. We surmise that this shortcoming is rooted in the evaluation of novel scenes, wherein the generalization capability of our model noticeably outperforms that of discriminative ones.

\subsection{Ablation Study}
We consider the test scene evaluation on the Replica dataset for the following ablation and exploration sections.

\label{sec:ablation}
\smallsec{Contributions of CTA and CVA:} We dissect the individual contributions of the proposed CTA and CVA modules in Table~\ref{tab:ablation}. An examination of the results reveals a more significant impact from the CTA module, when compared with the CVA module. We postulate that this occurs because the NeRF encoder and volume rendering have already mastered an implicit 3D representation, albeit falling short in modeling task relationships. Nevertheless, the integration of both modules yields further enhancement.

\smallsec{Choice of Condition NeRF:} We also ablate the conditional NeRF encoders with PixelNeRF~\cite{yu2021pixelnerf}, MVSNeRF~\cite{mvsnerf}, GNT~\cite{wang2022attention}, and GeoNeRF~\cite{geonerf2022} in Table~\ref{tab:nerf_encoder}. Note that we only adopt the encoder part defined in Section~\ref{sec:method}, causing a variation in performance from the full model. Each variant under our design surpasses its respective SS-NeRF baseline, affirming the universal advantage of our proposed CTA and CVA modules across different conditional NeRF encoders.

\smallsec{Effectiveness of CTA and CVA Modules:}
By integrating CVA and CTA modules, we concurrently increase the number of trainable parameters. To truly assess the advancements brought by the proposed CVA and CTA modules, we introduce an expanded version of SS-NeRF, designated as SS-NeRF (Enlarged). This variant features a doubled latent dimension, comprising 2.09M parameters and $6.18\times10^{11}$ FLOPs. The outcomes, as illustrated in Table~\ref{tab:enlarge}, confirm that MuvieNeRF's excellent performance is not simply the result of a more complex decoder, thereby highlighting the effectiveness of our module designs. Moreover, we further list the computational cost for our model and SS-NeRF in Table~\ref{tab:cost} to show that our CTA and CVA modules are indeed lightweight, but effective designs.

\begin{figure}[t]
		\centering
        \includegraphics[width =0.9\linewidth]{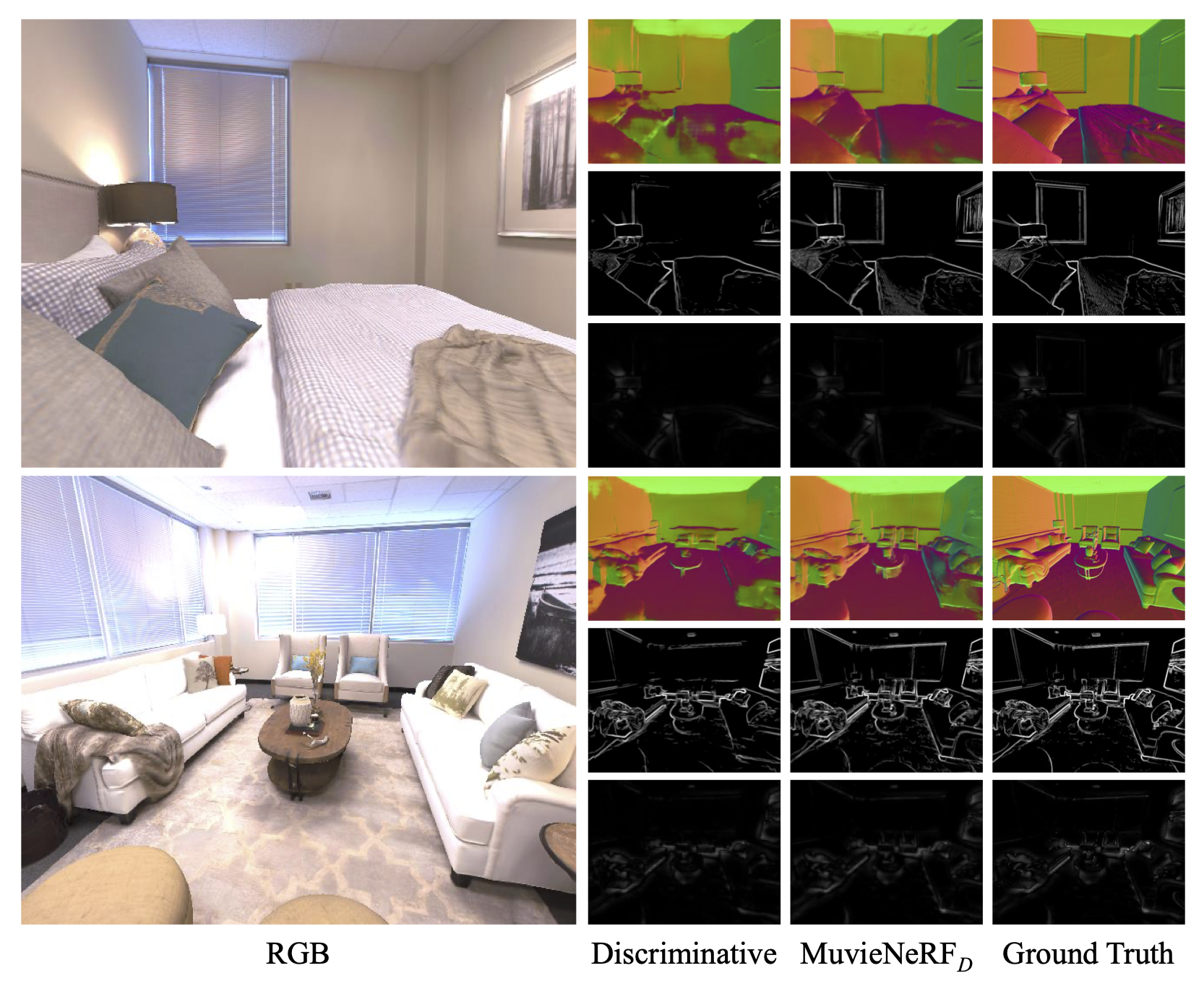}	
        \vspace{-5 pt}
		\caption{Visualizations for the setting without nearby-view annotations. our $\text{\modelname}_D$ markedly surpasses the discriminative baseline model~\cite{standley2020tasks} and can generate results closely paralleling the ground truth, indicating $\text{\modelname}_D$ is capable of tackling the more challenging settings.}
		\label{fig:challenging}
\end{figure}

\subsection{Additional Explorations}
\label{sec:exploration}

\subsubsection{Tackling Unknown Source-view Labels}
In this section, we apply the variant discussed in Section~\ref{sec:challenge} to tackle the more challenging problem setting with unknown nearby-view annotations. In Table~\ref{tab:challenging}, we show the quantitative comparisons for this variant, {\textbf{$\text{\modelname}_D$}}, and a hybrid baseline, \textbf{Discriminative}~\cite{standley2020tasks}, which shares almost the same architecture as our discriminative module. We use pre-trained weights from Taskonomy~\cite{Zamir_2018_CVPR} to initialize weights for both models. Limited by the computational constraint, we select three tasks, SN, ED, and KP with the closest relationships~\cite{standley2020tasks} for demonstration in this setting. 

In conjunction with the visualizations illustrated in Figure~\ref{fig:challenging}, it is clear that our $\text{\modelname}_D$ markedly surpasses the discriminative baseline model~\cite{standley2020tasks} and can generate results closely paralleling the ground truth. This observation indicates that our model is capable of tackling more challenging settings. We conjecture the reason is that the weighted sum format (Equation~\ref{eq:weightedsum}) enhances the fault tolerance of the predictions.

\begin{figure}[t]
		\centering
        \includegraphics[width =0.9\linewidth]{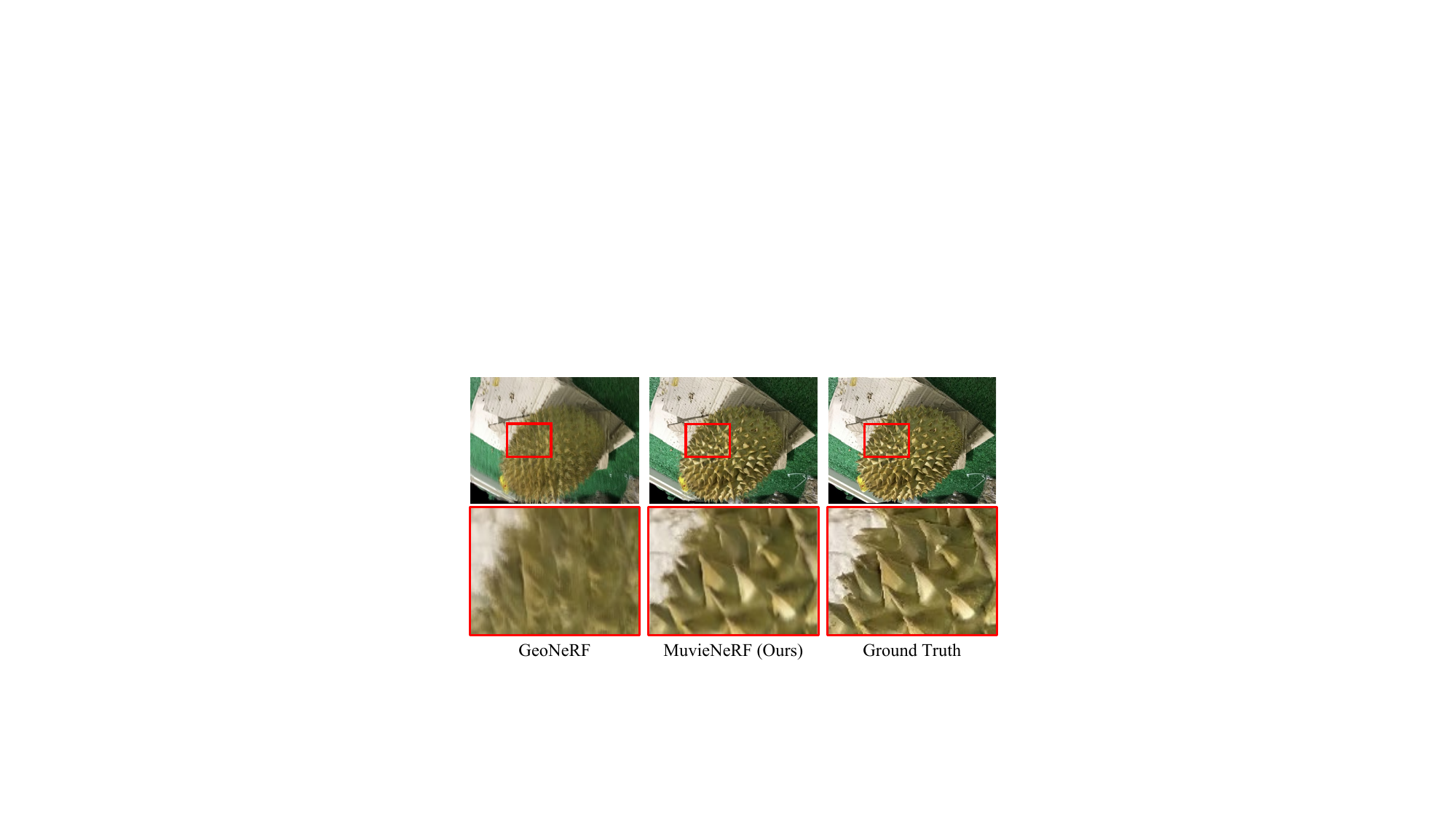}	
        \vspace{-2 pt}
		\caption{An out-of-distribution comparison on BlendedMVS dataset~\cite{yao2020blendedmvs} for GeoNeRF and our \modelname. Our model offers superior visual quality and preserves sharp contours. More visualizations about other OOD benchmarks are included in the appendix.
  }
  \vspace{-10pt}
		\label{fig:ood_comparison}
\end{figure}

\subsubsection{Out-of-distribution Generalization}

We demonstrate how the multi-task information learned from one dataset can effectively be utilized to benefit other datasets, by performing a zero-shot adaptation on out-of-distribution datasets with our \modelname trained on Replica. This takes a step further from investigating the generalization ability of unseen testing scenes in previous sections. We consider four datasets in total: LLFF~\cite{mildenhall2019local}, TartanAir~\cite{tartanair}, ScanNet~\cite{dai2017scannet}, and BlendedMVS~\cite{yao2020blendedmvs} containing indoor, outdoor, and even object-centric scenes. 

As an illustrative example, we showcase the comparison between the conditional NeRF backbone, GeoNeRF~\cite{geonerf2022}, and our \modelname for the novel-view RGB synthesis task in Table~\ref{tab:ood_rgb} and Figure~\ref{fig:ood_comparison}. Evidently, our model surpasses GeoNeRF by a significant margin, offering superior visual quality and retaining sharp contours, likely a result of the edge and surface normal information absorbed during the multi-task training. These outcomes substantiate that augmenting the model with more tasks, as part of our multi-task learning strategy, dramatically bolsters the generalization capability, thereby showcasing its immense potential for real-world applications.

\begin{table}[t]
\centering
    \resizebox{0.9\linewidth}{!}{
    \begin{tabular}{l|ccccccc}
Model & ScanNet & TartanAir & LLFF & BlendedMVS        \\ \hline
GeoNeRF & 31.71     & 26.51  & 20.68 &  16.27             \\ 
\modelname &  \textbf{32.76} &  \textbf{30.21}  &  \textbf{22.91}  &     \textbf{20.97}                 \\ 
\end{tabular}
}
\vspace{4pt}
    \caption{Averaged PSNR of out-of-distribution RGB synthesis task. Our model surpasses the GeoNeRF baseline by a large margin, affirming that the inclusion of additional tasks, as part of our approach, contributes to a substantial enhancement in the model's generalization capacity.
    }
    \vspace{-3pt}
    \label{tab:ood_rgb}
\end{table}

\subsection{Limitations and Future work}
\label{sec:limitation}
\smallsec{Limitations:} One major limitation of this work is the
reliance on data. \modelname requires images from dense
views, a requirement not fulfilled by most multi-task benchmarks. To circumvent this limitation, techniques that allow NeRF to learn from sparse views~\cite{Niemeyer2022Regnerf,zhang2021ners} could be adopted.

\smallsec{Task Relationships:} As elaborated in Section~\ref{sec:results}, SH and KP function as auxiliary tasks within the system. A deeper exploration into the relationships between tasks and the corresponding geometric underpinnings within our synthesis framework offers intriguing avenues for future research.

\smallsec{Extension to Other Synthesis Models:} 
We have demonstrated that incorporating cross-view geometry and cross-task knowledge can enhance multi-task learning for synthesis models. We anticipate that similar strategies could be extended to 3D synthesis models other than NeRF, such as point clouds~\cite{wiles2020synsin} and meshes~\cite{hu2021worldsheet,kong2023vmap}.

\section{Conclusion}
\label{sec:conclusion}

This paper introduces the novel concept of Multi-Task View Synthesis (MTVS), recasting multi-task learning as a set of view synthesis problems. Informed by this new perspective, we devise \modelname, a unified synthesis framework enriched with novel Cross-View Attention and Cross-Task Attention modules. \modelname enables the simultaneous synthesis of multiple scene properties from novel viewpoints.
Through extensive experimental evaluations, we establish \modelname's proficiency in addressing the MTVS task, with performance exceeding that of discriminative models across various settings. Our model also demonstrates broad applicability, extending to a variety of conditional NeRF backbones.

\section*{Acknowledgement}
We thank Kangle Deng, Derek Hoiem, Ziqi Pang, Deva Ramanan, Manolis Savva, Shen Zheng, Yuanyi Zhong, Jun-Yan Zhu, and Zhen Zhu for their valuable comments. 

This work was supported in part by NSF Grant 2106825, Toyota Research Institute, NIFA Award 2020-67021-32799, the Jump ARCHES endowment, the NCSA Fellows program, the Illinois-Insper Partnership, and the Amazon Research Award. This work used NVIDIA GPUs at NCSA Delta through allocations CIS220014 and CIS230012 from the ACCESS program.

{\small
\bibliographystyle{ieee_fullname}
\bibliography{egbib}
}

\clearpage
\newpage

\setcounter{figure}{0}
\setcounter{table}{0}
\renewcommand{\thefigure}{\Alph{figure}}
\renewcommand{\thetable}{\Alph{table}}

\appendix
In this appendix, we first provide additional qualitative results, including more visualizations of the two main datasets and results on other out-of-distribution datasets in Section~\ref{sec:more_results}. Next, we conduct additional experimental evaluations to analyze the behavior of our model under different settings in Section~\ref{sec:additionalexperiments}. We further present the multiple run results of our model and the compared methods in Section~\ref{sec:multiple_runs}, demonstrating that our \modelname~consistently achieves the best performance. Finally, we include additional details about our model implementation and dataset processing in Section~\ref{sec:implementation}.

\section{More Visualizations}
\label{sec:more_results}

We provide more qualitative results from the following two aspects: \textbf{(1)} visual comparison with other synthesis methods and \textbf{(2)} RGB synthesis results on other out-of-distribution datasets.

\subsection{Comparison with Other Synthesis Methods}

\begin{figure*}[t]
		\centering
        \includegraphics[width =\linewidth]{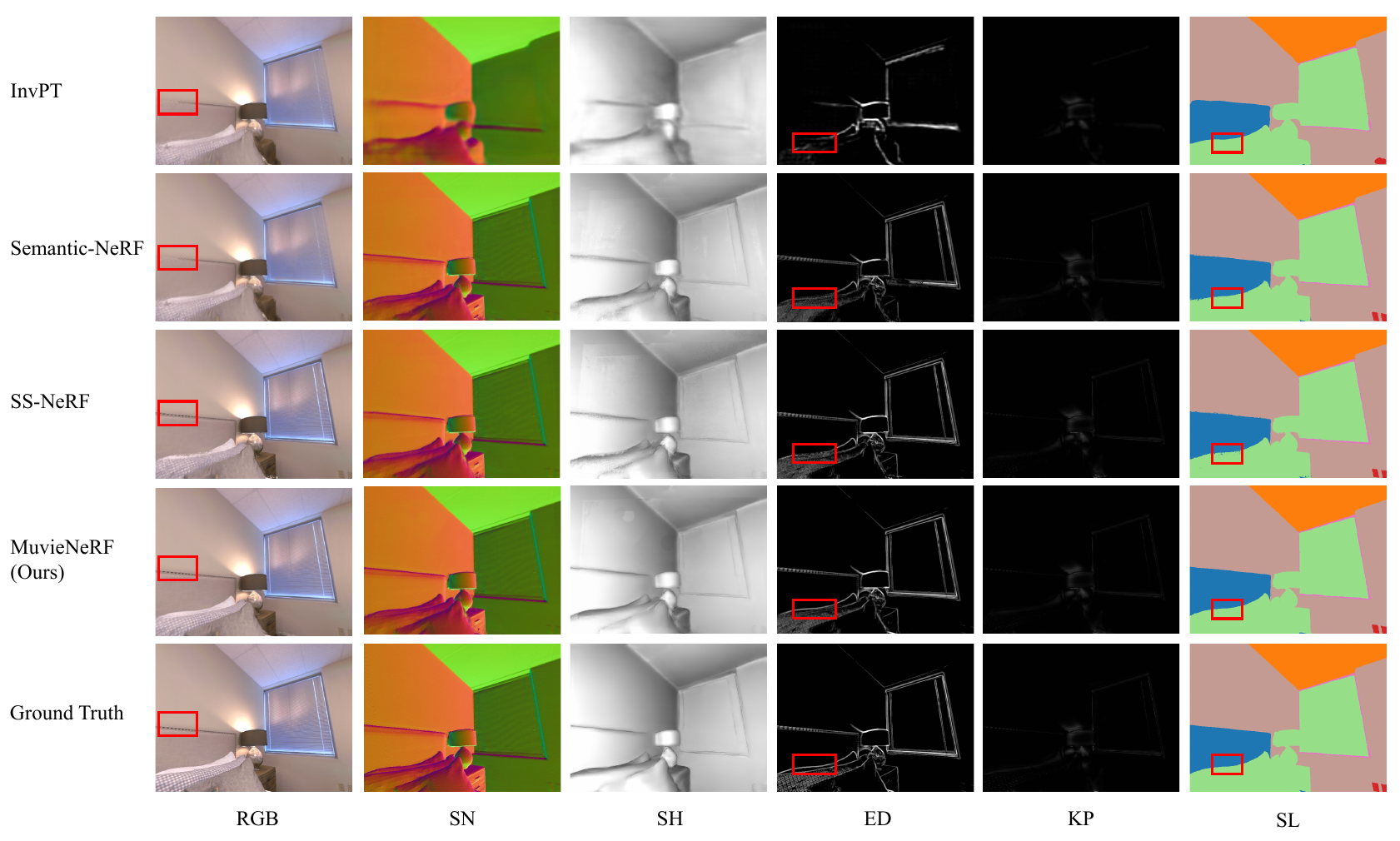}
		\caption{Additional qualitative results on one testing scene in the Replica dataset. Our proposed \modelname outperforms other methods with more accurate predictions and sharper boundaries, which demonstrates the effectiveness of the multi-task and cross-view information modeled by the CTA and CVA modules. \textbf{Zoom in to better see the comparison.}}
		\label{fig:additional_results}
\end{figure*}

\begin{figure*}[t]
		\centering
        \includegraphics[width =0.98\linewidth]{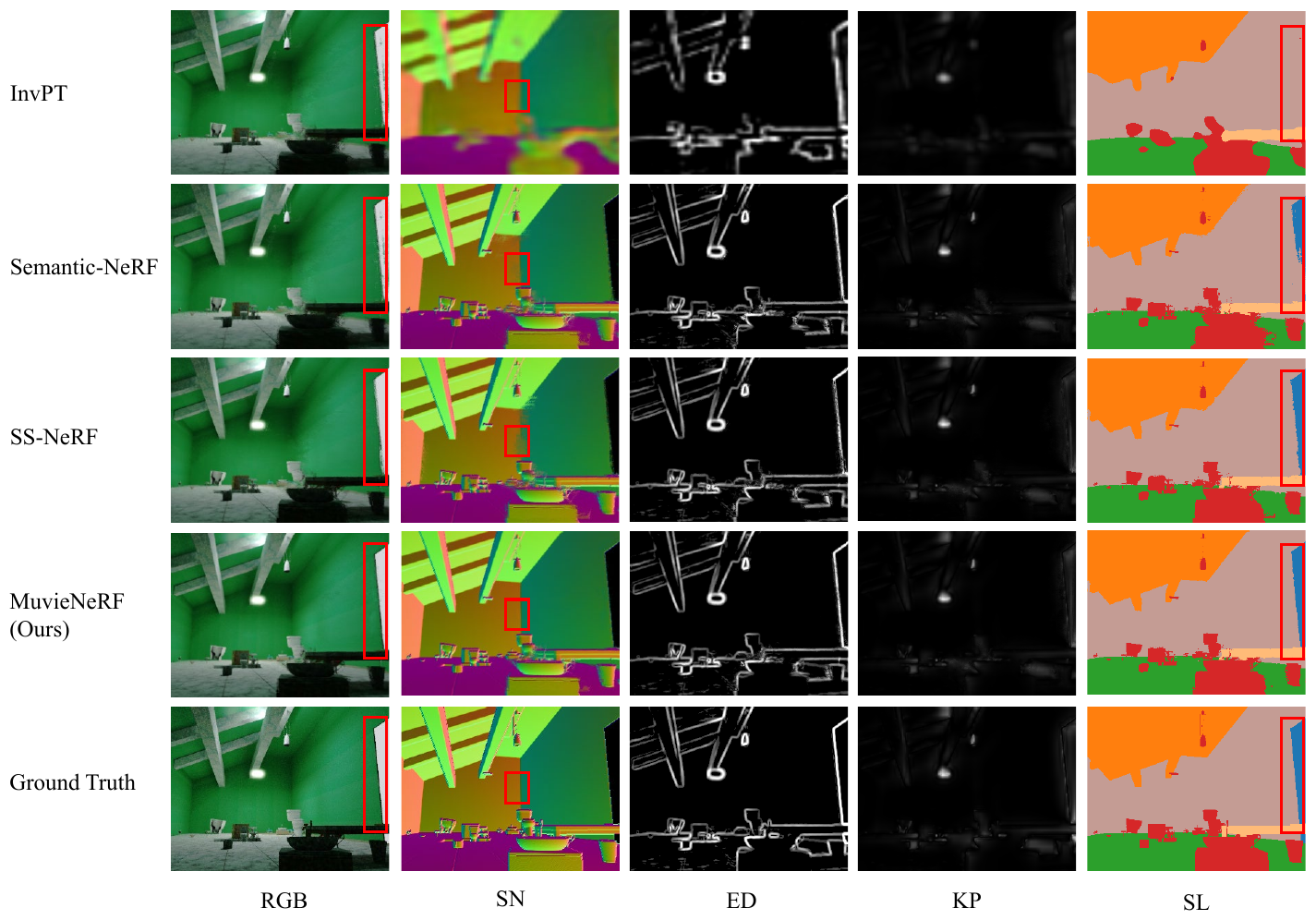}
		\caption{Additional qualitative results on one testing scene in the SceneNet RGB-D dataset. Our proposed \modelname outperforms other methods, indicating that our model benefits from the multi-task and cross-view information with the designed CTA and CVA modules. The black regions in the surface normal visualizations are due to the missing depth values in those regions. \textbf{Zoom in to better see the comparison.}}
		\label{fig:additional_results_scenenet}
\end{figure*}

Additional qualitative comparisons for all the compared methods in the Replica and SceneNet RGB-D datasets are shown in Figure~\ref{fig:additional_results} and Figure~\ref{fig:additional_results_scenenet}, respectively. Our \modelname~outperforms other methods with clearer and more accurate contours of the objects in scenes. This is because \modelname~utilizes the CTA and CVA modules to better take advantage of the shared knowledge across different downstream tasks and the cross-view information.

\subsection{Out-of-distribution Generalization}
In the main paper, we present a practical application of our proposed \modelname~to show that the multi-task information learned from one dataset can be generalized to the scenes in other datasets. We use \modelname~trained on the Replica dataset to perform a zero-shot adaption on out-of-distribution datasets: LLFF~\cite{mildenhall2019local}, TartanAir~\cite{tartanair}, ScanNet~\cite{dai2017scannet}, and BlendedMVS~\cite{yao2020blendedmvs} containing indoor, outdoor and even object-centric scenes. The detailed information of these four datasets is listed in Table~\ref{tab:datasets}.

\begin{figure}[t]
		\centering
        \includegraphics[width =\linewidth]{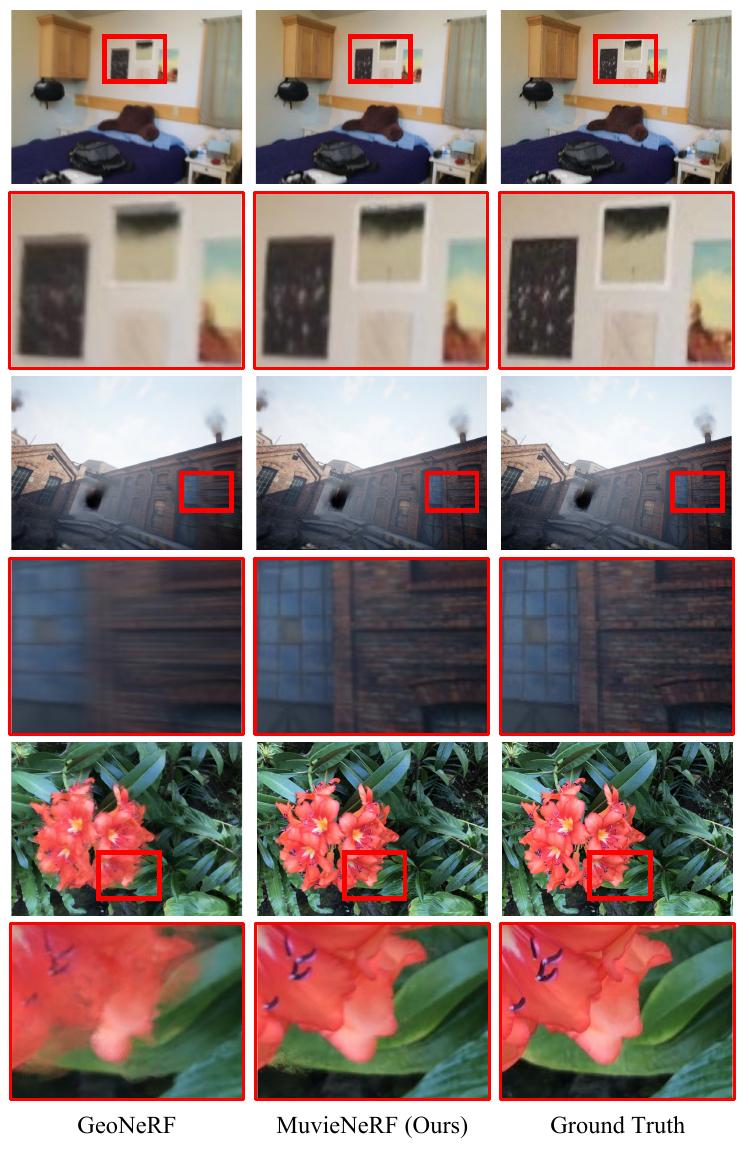}
    \caption{Additional qualitative RGB synthesis results on out-of-distribution datasets. \textbf{From top to bottom:} ScanNet~\cite{dai2017scannet}, TartanAir~\cite{tartanair}, and LLFF~\cite{mildenhall2019local}. Our \modelname~yields better visual quality, demonstrating that the multi-task and cross-view knowledge learned during training can be generalized and applied to out-of-distribution datasets. \textbf{Zoom in to better see the comparison.}}
		\label{fig:out_of_distribution}
\end{figure}

The RGB synthesis results on those out-of-distribution datasets are shown in Figure~\ref{fig:out_of_distribution}. We can observe that our model renders higher-quality RGB images from novel views with sharper contours compared to the GeoNeRF baseline. The underlying reason lies in the joint modeling of edges and surface normal during training, which makes RGB prediction more precise even for the out-of-distribution datasets.

\begin{table}[t]
\centering
    \resizebox{\linewidth}{!}{
    \begin{tabular}{l|c|c|c|cccc}
Datasets & ScanNet~\cite{dai2017scannet} & TartanAir~\cite{tartanair} & LLFF~\cite{mildenhall2019local} & BlendedMVS~\cite{yao2020blendedmvs}        \\ \hline
Number of scenes & 4     & 4  & 8 & 2             \\  
Resolution &  384$\times$288  &  640$\times$480 & 1008$\times$756  &  768$\times$576                    \\  
Contents  & Indoor  & Indoor, Outdoor  & Indoor, Outdoor, Object  & Object
\end{tabular}
}
\vspace{4pt}
    \caption{Detailed information about the four out-of-distribution datasets, which contain indoor, outdoor, and/or even object-centric scenes.}
    \label{tab:datasets}
\end{table}

\begin{table}[t]
\centering
    \resizebox{\linewidth}{!}{
    \begin{tabular}{c|c|ccccccc}
Index   & Training Scene Name& SN  & SH  & ED & KP & SL     \\ \hline
1    & apartment\_0 (a) & $\times$   & $\times$    & $\checkmark$ & $\times$ & $\checkmark$     \\ 
2    & apartment\_0 (b) & $\checkmark$   & $\times$    & $\checkmark$ & $\checkmark$ & $\checkmark$     \\ 
3    & apartment\_1 & $\checkmark$   & $\checkmark$    & $\checkmark$ & $\checkmark$ & $\times$     \\ 
4    & apartment\_2 (a) & $\times$   & $\checkmark$    & $\checkmark$ & $\checkmark$ & $\checkmark$     \\ 
5    & apartment\_2 (b) & $\checkmark$   & $\checkmark$    & $\checkmark$ & $\checkmark$ & $\checkmark$     \\ 
6    & apartment\_2 (c) & $\times$   & $\checkmark$    & $\times$ & $\checkmark$ & $\times$     \\ 
7    & FRL\_apartment\_0 (a) & $\checkmark$   & $\checkmark$    & $\checkmark$ & $\checkmark$ & $\checkmark$     \\
8    & FRL\_apartment\_0 (b) & $\checkmark$   & $\times$    & $\times$ & $\checkmark$ & $\checkmark$     \\ 
9    & hotel\_0 (a) & $\times$   & $\checkmark$    & $\checkmark$ & $\checkmark$ & $\times$     \\ 
10    & hotel\_0 (b) & $\checkmark$   & $\checkmark$    & $\checkmark$ & $\times$ & $\times$     \\ 
11    & hotel\_0 (c) & $\times$   & $\checkmark$    & $\checkmark$ & $\times$ & $\checkmark$     \\ 
12    & hotel\_0 (d) & $\checkmark$   & $\checkmark$    & $\times$ & $\checkmark$ & $\checkmark$     \\ 
13    & office\_0 (a) & $\checkmark$   & $\times$    & $\times$ & $\times$ & $\checkmark$     \\ 
14    & office\_0 (b) & $\times$   & $\checkmark$    & $\checkmark$ & $\times$ & $\checkmark$     \\ 
15    & office\_0 (c) & $\checkmark$   & $\checkmark$    & $\times$ & $\checkmark$ & $\times$     \\ 
16    & office\_2 & $\times$   & $\checkmark$    & $\checkmark$ & $\checkmark$ & $\checkmark$     \\ 
17    & room\_2 (a) & $\checkmark$   & $\checkmark$    & $\checkmark$ & $\times$ & $\times$     \\ 
18   & room\_2 (b) & $\checkmark$   & $\times$    & $\times$ & $\checkmark$ & $\checkmark$     \\
\end{tabular}
}
\vspace{4pt}
    \caption{Simulated {\em federated training} setting where some of the task annotations for certain training scenes are unavailable.}
    \label{tab:federal_dataset}
\end{table}

\section{Additional Experimental Evaluation}
\label{sec:additionalexperiments}

We provide additional experimental evaluations to understand the behavior of \modelname and its capability from various aspects: (1) we evaluate our model in a {\em federated training} setting; (2) we ablate the CTA module with a lightweight choice of the cross-stitch module~\cite{misra2016cross}; (3) we report the results with a half-sized training set; (4) we provide an additional comparison for the discriminative models with extra data;  and (6) we ablate the contributions of the proposed CTA and CVA modules in the more challenging setting formulated by Equation~\ref{eq:challenging}.

\subsection{Federated Training with Partial Annotations}
\label{sec:limited}

In the real-world regime, it is not always possible to get access to all the different types of annotations to train a model. In this scenario, federated training~\cite{federated2020} is widely used. To simulate the real-world regime, we propose such a setting where {\em every task annotation for each training scene has a 30\% probability of being unavailable}. The detailed setting is shown in Table~\ref{tab:federal_dataset}, where only 2 scenes get access to all annotations. We train our \modelname~on this subset and compare against the two NeRF-based baselines Semantic-NeRF~\cite{semantic_nerf} and SS-NeRF~\cite{ssnerf2023} trained on the full training set. The evaluation is conducted on the same testing scenes in Replica.

The results are shown in Table~\ref{tab:ab_federal}. We have the following observations. First, our model still outperforms the two baselines even with missing annotations, indicating that leveraging multi-task and cross-view information in our proposed \modelname~is the key to the success. Second, we still achieve comparable results to the model trained with the full training set, showing the robustness and label-efficiency of our method.

\begin{table}[t]
\centering
    \resizebox{\linewidth}{!}{
    \begin{tabular}{l|cccccc}
Settings      & RGB ($\uparrow$)                    & SN ($\downarrow$) & SH ($\downarrow$) & ED ($\downarrow$) & KP ($\downarrow$) & SL ($\uparrow$)      \\ \hline
Full set + Semantic-NeRF & 27.08 & 0.0221 & 0.0418 & 0.0212 & 0.0055 & 0.9417                     \\
Full set + SS-NeRF      & 27.22 & 0.0224 & 0.0405 & 0.0196 & 0.0053 & 0.9483  \\ 
Full set + \modelname & 28.55 & 0.0201 & 0.0408 & 0.0162 & 0.0051 & 0.9563                       \\ \hline
w/o Full set + \modelname & 27.86 & 0.0212 & 0.0422 & 0.0185 & 0.0053 & 0.9526               \\ 
\end{tabular}
}
\vspace{4pt}
    \caption{Comparison between our model and two baselines in a federated training setting. Our \modelname~model still outperforms the two baselines and even achieves comparable results to the model trained with the full set, indicating the effectiveness and robustness of the proposed method.}
    \label{tab:ab_federal}
\end{table}

\subsection{Comparison with Lightweight Cross-task Modules}

The novel CVA and CTA modules are designed to facilitate multi-task and cross-view information interaction, which improves the performance of \modelname. In this section, we provide an additional ablation with a lightweight choice of CTA modules. We show in the following that although the simpler module reaches comparable performance when modeling RGB together with two additional tasks, its performance significantly lags behind our novel design when handling the more challenging setting with RGB modeled with five additional tasks. It demonstrates that the designed CVA and CTA modules in our \modelname~have a larger capacity for modeling multiple tasks.

Concretely, we adopt the cross-stitch~\cite{misra2016cross} module for experimental evaluation. The cross-stitch module takes a simple strategy of performing a learned combination of task-specific features. More specifically, when applied in our \modelname~pipeline, it functions after the ``separate decoders" as
\begin{equation}
F_{\rm out} = \mathbf{W}F_{\rm in},
     \label{eq:feature_interaction_arbitrary}
\end{equation}
where $F_{\rm in}, F_{\rm out} \in \mathbb{R}^{K \times V \times c}$ are the input and output of the cross-stitch module, respectively. $\mathbf{W} \in \mathbb{R}^{K \times K}$ is a learnable weight matrix with an $L_2$ regularization for each row. Each weight value $w_{ij}$ measures the information the $j$-th component obtained from the $i$-th component.

The experimental comparison of our \modelname~and the simpler cross-stitch implementation is shown in Table~\ref{tab:capacity}. When modeling with only two additional tasks, the cross-stitch module could reach comparable performance to our method. However, when the number of tasks jointly learned with RGB increases to five, the cross-stitch implementation fails to serve as an efficient multi-task learning strategy. This indicates that although the simpler cross-stitch module can afford to benefit the information exchange in the easier cases when two tasks beyond RGB are jointly modeled, it does not have enough capacity to handle the more complicated relationships of five tasks along with RGB. In comparison, our design of the CVA and CTA modules is superior, which leads to the success of modeling more tasks.

\begin{table*}[t]
\centering
    \resizebox{\linewidth}{!}{
    \begin{tabular}{l|ccccc|ccccc|ccccc}
     \multirow{2}{*}{Model} & \multicolumn{5}{c|}{NeRF's Images (No Tuned)} & \multicolumn{5}{c|}{NeRF's Images (Tuned)} & \multicolumn{5}{c}{GT Images (Upper Bound)} \\     \cline{2-16} 
      & SN ($\downarrow$) & SH ($\downarrow$) & ED ($\downarrow$) & KP ($\downarrow$) & SL ($\uparrow$)                   & SN ($\downarrow$) & SH ($\downarrow$) & ED ($\downarrow$) & KP ($\downarrow$) & SL ($\uparrow$) & SN ($\downarrow$) & SH ($\downarrow$) & ED ($\downarrow$) & KP ($\downarrow$) & SL ($\uparrow$)\\ \hline
    Taskgrouping (15k) & 0.0464 & 0.0757 & 0.0418 & 0.0088 &  0.6633 &0.0479 & 0.0531 &0.0388  &0.0087  & 0.7193 & 0.0438 & 0.0509 & 0.0284 & 0.0058 & 0.7509 \\ 
      MTI-Net (15k) & 0.0533 & 0.0676 & 0.0414 & 0.0089 & 0.5509 & 0.0463 &  0.0581 & 0.0314 & 0.0079 &  0.6821 & 0.0462 & 0.0500 & 0.0271 & 0.0050 & 0.7555 \\
      InvPT (15k) & 0.0463 & 0.0580 & 0.0417 & 0.0079 & 0.7157 & 0.0399 & 0.0477 & 0.0272 & 0.0057 & 0.7719 & 0.0402 & 0.0472 & 0.0257 & 0.0047 & 0.7981 \\ \hline
      Taskgrouping-4M &  0.0451 & - & 0.0350 & 0.0079 & 0.6692 &  0.0313 & - &  0.0311 &  0.0066&  0.7818 &  0.0231 & - & 0.0112 & 0.0040 & 0.8376 \\ \hline
     MuvieNeRF & \bf 0.0201 & \bf 0.0408 & \bf 0.0162 & \bf 0.0051 & \bf 0.9563 & - & - & - & - & - & - & - & - & - & -
\end{tabular}
}
\vspace{4pt}
    \caption{Additional comparison with discriminative models. Training discriminative models with a larger amount of data still cannot outperform our MuvieNeRF, indicating that the discriminative models still lack the ability of multi-view reasoning even when the training data increases.
    }
    \label{suptab:discriminative}
\end{table*}

\begin{table}[t]
\centering
    \resizebox{0.9\linewidth}{!}{
    \begin{tabular}{l|ccc}
Tasks & RGB ($\uparrow$)                                    & SN ($\downarrow$)  & SL ($\uparrow$) \\ \hline
Cross-stitch (RGB + 2 Tasks) & \textbf{27.16}    &   0.0242   &    \textbf{0.9519}             \\ 
\modelname~(RGB + 2 Tasks) & 26.97  & \textbf{0.0229}  &     0.9476                       \\ \hline
Cross-stitch (RGB + 5 Tasks) &  27.57  & 0.0219  &  0.9459                  \\ 
\modelname~(RGB + 5 Tasks)&  \textbf{28.55}    & \textbf{0.0201} &    \textbf{0.9563}                          \\ 
 
\end{tabular}
}
\vspace{4pt}
    \caption{Comparison between our \modelname~model design and a simpler cross-stitch~\cite{misra2016cross} multi-task module. The results are averaged over the testing scenes on the Replica dataset. The simpler cross-stitch implementation can reach comparable results when the target is easier (RGB + 2 tasks), but fails to achieve satisfactory results when the target becomes more challenging (RGB + 5 tasks). In comparison, our model is able to achieve better performance with more tasks learned together.}
    \label{tab:capacity}
\end{table}

\subsection{Results with a Half-sized Training Set}

We further investigate the robustness of our model by decreasing the number of scenes in the training dataset to only half of the original size. The results are shown in Table~\ref{tab:ab_training_size}. We could observe a similar phenomenon as the federated training result in Table~\ref{tab:ab_federal}: when the number of training scenes reduces, the performance of our model only drops slightly while still outperforming the other compared methods, demonstrating the robustness and sample-efficiency of our method.

\subsection{Additional Comparisons with Discriminative Models}

We add the following two sets of comparisons for discriminative models on Table~\ref{suptab:discriminative}: (1) we use 15K images rendered from the Replica dataset for training; (2) we use a pre-trained checkpoint (Taskgrouping-4M, the only available multi-task one with 4 tasks) on Taskonomy ($\sim${4M} data) for initialization and finetune it on Replica. All these variants still cannot outperform our MuvieNeRF, indicating that the discriminative models still \textbf{lack the ability of multi-view reasoning} even when the training data increases.

\begin{table}[t]
\centering
    \resizebox{\linewidth}{!}{
    \begin{tabular}{l|cccccc}
Settings      & RGB ($\uparrow$)                    & SN ($\downarrow$) & SH ($\downarrow$) & ED ($\downarrow$) & KP ($\downarrow$) & SL ($\uparrow$)      \\ \hline
Full + Semantic-NeRF & 27.08 & 0.0221 & 0.0418 & 0.0212 & 0.0055 & 0.9417                     \\
Full + SS-NeRF & 27.22 & 0.0224 & 0.0405 & 0.0196 & 0.0053 & 0.9483   \\ 
Full + \modelname & 28.55 & 0.0201 & 0.0408 & 0.0162 & 0.0051 & 0.9563              \\ \hline
Half + \modelname & 28.11 & 0.0211 & 0.0427 & 0.0168 & 0.0054 & 0.9562               \\ 
\end{tabular}
}
\vspace{4pt}
    \caption{Comparison of training with only half training scenes in Replica. Our model still achieves relatively satisfactory results when the number of training scenes reduces to only half, indicating the sample-efficiency and robustness of our method.}
    \label{tab:ab_training_size}
\end{table}

\begin{table*}[t]
\centering
    \resizebox{\linewidth}{!}{
    \begin{tabular}{c|l|cccccc}

    \multicolumn{2}{c|}{Tasks} & RGB ($\uparrow$)                    & SN ($\downarrow$) & SH ($\downarrow$) & ED ($\downarrow$) & KP ($\downarrow$) & SL ($\uparrow$)      \\ \hline
    \multirow{3}{*}{\shortstack{Training \\ scene \\ evaluation}} & Semantic-NeRF & \multicolumn{1}{c}{33.79 ($\pm$0.1579)}     & \multicolumn{1}{c}{0.0231 ($\pm$0.0013)}     &    0.0400 ($\pm$0.0005)   &         0.0127 ($\pm$0.0003)             &           0.0037 ($\pm$0.0000)          &           0.9522 ($\pm$0.0017)                        \\ 
    & SS-NeRF & \multicolumn{1}{c}{34.07 ($\pm$0.2572)}     & \multicolumn{1}{c}{0.0212 ($\pm$0.0008)}     &           0.0379 ($\pm$0.0007)               &      0.0113 ($\pm$0.0005)               &             0.0035 ($\pm$0.0000)          &               0.9528 ($\pm$0.0023)   \\ 
    & \modelname & \textbf{34.85 ($\pm$0.1440)} & \textbf{0.0197 ($\pm$0.0003)} & \textbf{0.0352 ($\pm$0.0006)} & \textbf{0.0102 ($\pm$0.0003)} & \textbf{0.0034 ($\pm$0.0000)} & \textbf{0.9589 ($\pm$0.0009)}             \\ \hline
    \multirow{3}{*}{\shortstack{Testing \\ scene \\ evaluation}} & Semantic-NeRF & \multicolumn{1}{c}{26.94 ($\pm$0.3180)}     & \multicolumn{1}{c}{0.0219 ($\pm$0.0004)}     &    0.0410 ($\pm$0.0005)   &         0.0195 ($\pm$0.0018)             &           0.0054 ($\pm$0.0001)          &           0.9502 ($\pm$0.0053)                        \\
    & SS-NeRF & \multicolumn{1}{c}{27.65 ($\pm$0.6055)}     & \multicolumn{1}{c}{0.0216 ($\pm$0.0010)}     &           0.0405 ($\pm$0.0004)               &      0.0184 ($\pm$0.0016)               &             0.0053 ($\pm$0.0001)          &               0.9503 ($\pm$0.0070)   \\ 
    & \modelname & \textbf{28.50 ($\pm$0.2127)} & \textbf{0.0200 ($\pm$0.0002)} & \textbf{0.0402 ($\pm$0.0006)} & \textbf{0.0164 ($\pm$0.0004)} & \textbf{0.0051 ($\pm$0.0001)} & \textbf{0.9586 ($\pm$0.0033)}              \\ 
\end{tabular}
}
\vspace{4pt}
    \caption{Results of all the compared models with four multiple runs on the Replica dataset. Our \modelname~consistently has better performance and overall smaller deviation among multiple runs than the single-task Semantic-NeRF~\cite{semantic_nerf} and the multi-task SS-NeRF~\cite{ssnerf2023}, demonstrating the effectiveness of our model design.}
    \label{tab:multiple_runs}
\end{table*}

\subsection{Contributions of CTA and CVA Modules with the More Challenging Setting}

In Table~\ref{tab:ablation} in the main paper, we dissect the individual contributions of the proposed CTA and CVA modules with our primary setting. We additionally ablate their contributions of them in the more challenging setting formulated by Equation~\ref{eq:challenging}. The results in Table~\ref{suptab:ablation} show similar conclusions and validate the proposed CVA and CTA modules are universally beneficial.

\begin{table}[t]
\centering
    \resizebox{0.7\linewidth}{!}{
    \begin{tabular}{l|ccc}
    Model &  SN ($\downarrow$) &  ED ($\downarrow$) & KP ($\downarrow$) \\ \hline
    $\text{\modelname}_\mathrm{w/o~CTA}$ &  0.0694 &  0.0256 & 0.0079 \\
      $\text{\modelname}_\mathrm{w/o~CVA}$ & 0.0668 &  0.0246 & 0.0076 \\
      $\text{\modelname}_D$ & \bf 0.0605 & \bf 0.0230 & \bf 0.0074\\
\end{tabular}
}
\vspace{4pt}
    \caption{Ablation study with CTA and CVA modules on  Replica~\cite{straub2019replica} dataset with the more challenging setting. $\text{\modelname}_\mathrm{w/o~CTA}$ is the variant without CTA module; $\text{\modelname}_\mathrm{w/o~CVA}$ is the variant without CVA module. The proposed CVA and CTA modules are universally beneficial for both problem settings.}
    \label{suptab:ablation}
\end{table}

\section{Multiple Runs}
\label{sec:multiple_runs}

To further validate the robustness and good performance of our model against other methods, we show the results of multiple runs on the Replica dataset in Table~\ref{tab:multiple_runs}. Our \modelname~consistently outperforms the single-task Semantic-NeRF~\cite{semantic_nerf} and the multi-task SS-NeRF~\cite{ssnerf2023} baselines, demonstrating the effectiveness of our model design.

\section{Implementation Details}
\label{sec:implementation}

We provide the architecture of the conditional NeRF encoders and the additional U-Net~\cite{ronneberger2015u} discriminative module we used for Section~4.5. More details of the training procedure and dataset processing are also included.

\subsection{Conditional NeRF Encoders}
\label{sec:nerfencoder}

\smallsec{GeoNeRF~\cite{geonerf2022} encoder} first uses a feature pyramid network~\cite{fpn2017} to encode input views of the scene to cascaded cost volumes~\cite{gu2020cascade}. Next, it masks out the input view features when the depth of the current 3D point is larger than the estimated depth in the corresponding input view. Finally, four cross-view attention operations are used to process the multi-view tokens. We refer to the official repository~\footnote{\url{https://github.com/idiap/GeoNeRF}} of GeoNeRF for our implementation.

\smallsec{MVSNeRF~\cite{mvsnerf} encoder} takes a similar architecture to the GeoNeRF encoder only without the cross-view attention modules. We refer to the released codes~\footnote{\url{https://github.com/apchenstu/mvsnerf}} for implementation.

\smallsec{PixelNeRF~\cite{yu2021pixelnerf} encoder} uses ResNet-34~\cite{resnet16} as the backbone of its feature extractor. It chooses the features prior to the first four pooling layers and upsamples them to be in the same shape as the input RGB images to obtain the multi-scale features. Next, the sampled points are projected to the image planes of the input views to obtain the projected feature from the $V$ source views. We implement it based on the official repository~\footnote{\url{https://github.com/sxyu/pixel-nerf}}.

\smallsec{GNT~\cite{wang2022attention} encoder}~\footnote{\url{https://github.com/VITA-Group/GNT}} also adopts ResNet-34 as the feature encoder to obtain the multi-view features from multi-view RGB inputs. We apply the same strategy as the PixelNeRF encoder to obtain the features for single 3D points. Notice that, in the original GNT model which is solely designed for RGB synthesis, the multi-view features further go through a view transformer~\cite{attention17}. However, the output of their transformer is not compatible with our designed decoder pipeline so we only treat the ResNet part as the encoder. Therefore, the GNT encoder serves as the single-scale version of the PixelNeRF encoder in our experiments and it can explain the reason why the GNT encoder performs the worst in our main paper.

\subsection{The Additional Discriminative Module}

In Sections~3.4 and 4.5 we introduce the model \modelname$_{D}$ for the more challenging problem setting with unknown nearby-view annotations. We take the encoder-decoder structure used in~\cite{standley2020tasks} for the U-Net shaped module $F_{\rm UNet}$, which takes RGB images as the input and predicts pixel-level scene properties.

Concretely, for the U-Net module, we use a shared encoder with the Xception~\cite{xception} as the backbone and apply $K$ light-weighted deconvolutional layers~\cite{deconv} to predict multiple scene properties. After the predictions, we use the 3D coordinate of the queried point to project the sampled points to the input image planes to obtain the single-pixel scene properties for the weighted sum.

\subsection{Training Details}

We set the weights for the six chosen tasks as $\lambda_{{\rm RGB}}=1$, $\lambda_{{\rm SN}}=1$, $\lambda_{\rm SL}=0.04$, $\lambda_{\rm SH}=0.1$, $\lambda_{\rm KP}=2$, and $\lambda_{\rm ED}=0.4$ based on empirical observations. We use the Adam~\cite{kingma2014adam} optimizer with an initial learning rate of $5\times 10^{-4}$ and set $\beta_{1}=0.9, \beta_{2}=0.999$. During training, each iteration contains a batch size of 1024 rays randomly sampled from all training scenes. The number of input views is set to 5. Following~\cite{misra2016cross}, we adopt a two-stage training strategy. We first train all the parameters except for the self-attention modules in the CTA module for $5\times 10^{3}$ iterations. Afterwards, we train the parameters in the self-attention modules along with other parameters for $1\times 10^{3}$ iterations. We train our model on a single NVIDIA A100 with 40GB memory for around 2.5 hours.

\subsection{Datasets Details}

\smallsec{Replica dataset~\cite{straub2019replica}} is a synthetic dataset which has accurate 3D mesh, semantic annotations and depth information. For semantic labels (SL), we map the original 88-class semantic labels in Replica dataset to the commonly-used 13-class annotation defined in NYUv2-13~\cite{Silberman:ECCV12}. For surface normal (SN), we derive it from depth:
\begin{equation}
    \textrm{SN}(x,y,z) = (- \frac{dz}{dx}, -\frac{dz}{dy},1),
\end{equation}
where ($x,y,z$) is the \emph{3D} coordinate and $\frac{dz}{dx}$, $\frac{dz}{dy}$ are the gradients of $x$ and $y$ with respect to $z$, respectively. Edge (ED) and keypoint (KP) are rendered with Canny~\cite{canny} edge detector and SIFT~\cite{sift2004}. Shadings (SH) are obtained by XTConsistency~\cite{zamir2020robust} which are pre-trained on indoor scenes. To better satisfy the multi-task setting in the real world with unknown camera poses, we generate the poses of each scene with COLMAP~\cite{schoenberger2016sfm}.

\smallsec{SceneNet RGB-D dataset~\cite{mccormac2016scenenet}} is a large-scale photorealistic dataset that allows rendering RGB images along with pixel-wise semantic and depth annotations. We use the same strategy as the Replica dataset to obtain the semantic labels and surface normal for SceneNet RGB-D. We also use Canny and SIFT to render the ED and KP annotations. The pre-trained model for SH failed to work on this dataset; therefore, we discard shadings for the evaluation on SceneNet RGB-D.

\end{document}